\documentclass[12pt]{article}

    \usepackage[utf8]{inputenc}
    \usepackage[
        twoside,
        top=1in,
        bottom=0.75in,
        inner=0.75in,
        outer=0.75in
    ]{geometry}
    \usepackage[square, numbers]{natbib}
    \usepackage{color}

    \usepackage{amsfonts,amsmath,amssymb}
    \usepackage{titling}

    \usepackage{graphicx}
    \usepackage{hyperref}
    
    \usepackage{caption}
    \usepackage{subcaption}

    \usepackage[export]{adjustbox}

    \usepackage{multirow}
    \usepackage{tcolorbox}
    
    \usepackage{listings,xcolor}

    \usepackage{booktabs} 

    \usepackage{authblk}

\colorlet{punct}{gray!60!black}
\definecolor{background}{HTML}{f1f1f1}
\definecolor{delim}{HTML}{999999}
\definecolor{eclipseStrings}{HTML}{333333}
\definecolor{eclipseKeywords}{HTML}{444444}
\colorlet{numb}{gray!60!black}
\lstdefinelanguage{json}{
    basicstyle=\tiny, 
    stringstyle=\color{eclipseKeywords}, 
    numbers=left,
    numberstyle=\scriptsize,
    stepnumber=1,
    numbersep=8pt,
    showstringspaces=false,
    breaklines=true,
    backgroundcolor=\color{background},
    string=[s]{"}{"},
    comment=[l]{:\ "},
    morecomment=[l]{:"},
    literate=
     *{0}{{{\color{numb}0}}}{1}
      {1}{{{\color{numb}1}}}{1}
      {2}{{{\color{numb}2}}}{1}
      {3}{{{\color{numb}3}}}{1}
      {4}{{{\color{numb}4}}}{1}
      {5}{{{\color{numb}5}}}{1}
      {6}{{{\color{numb}6}}}{1}
      {7}{{{\color{numb}7}}}{1}
      {8}{{{\color{numb}8}}}{1}
      {9}{{{\color{numb}9}}}{1}
      {:}{{{\color{punct}{:}}}}{1}
      {,}{{{\color{punct}{,}}}}{1}
      {\{}{{{\color{delim}{\{}}}}{1}
      {\}}{{{\color{delim}{\}}}}}{1}
      {[}{{{\color{delim}{[}}}}{1}
      {]}{{{\color{delim}{]}}}}{1},
}

\definecolor{promptdelim}{HTML}{4aa567}
\lstdefinelanguage{text}{
    basicstyle=\tiny, 
    stringstyle=\color{eclipseKeywords}, 
    breaklines=true,
    breakindent=0pt,
    framextopmargin=5pt,
    framexbottommargin=5pt,
    framexleftmargin=3pt,
    framexrightmargin=3pt,
    backgroundcolor=\color{background},
    literate=
        {\{}{{{\color{promptdelim}{\{}}}}{1}
        {\}}{{{\color{promptdelim}{\}}}}}{1},
}

\newcommand{\storyscript}[2]{
  \begin{tcolorbox}[
    beforeafter skip=5pt,
    halign=flush center,
    colframe=white,
    #2,
  ]
  \color{black}\small #1
  \end{tcolorbox}
}

\newcommand{\tightscript}[2]{
  \begin{tcolorbox}[
    beforeafter skip=5pt,
    left=1mm,right=1mm,top=2mm,bottom=2mm,
    halign=flush center,
    colframe=white,
    #2,
  ]
  \color{black}\small #1
  \end{tcolorbox}
}

\usepackage{varwidth}
\usepackage{environ}
\usepackage{xparse}
\usepackage{tcolorbox}
\tcbuselibrary{skins}

\newlength{\bubblewidth}
\AtBeginDocument{\setlength{\bubblewidth}{.75\linewidth}}
\definecolor{bubblegreen}{HTML}{c6e5ba}
\definecolor{bubbledarkgreen}{RGB}{  97, 121, 107  }
\definecolor{bubblegray}{HTML}{f1f1f1}
\definecolor{usernamecolor}{gray}{0.5}

\newcommand{\bubble}[4]{%
  \noindent \textcolor{usernamecolor}{\small #1}
  \nopagebreak
  \tcbox[
    beforeafter skip=2pt,
    left=1mm,right=1mm,top=0mm,bottom=0mm,
    colback=#2,
    colframe=#2,
    #4,
  ]{\color{black}\begin{varwidth}{\bubblewidth}\small #3\end{varwidth}}%
}
\newcommand{\bubbleat}[5]{%
  \noindent \textcolor{usernamecolor}{\small #1}
  \nopagebreak
  \tcbox[beforeafter skip=2pt,left=1mm,right=1mm,top=0mm,bottom=0mm,colback=#2,colframe=#2,#4,]{\color{black}
  \begin{varwidth}{\bubblewidth}
   {\color{bubbledarkgreen}\small@#5} \small#3
  \end{varwidth}}%
}
\newcommand{\rightbubbleat}[4]{
  {\raggedleft
  \bubbleat{#1}{#2}{#3}{sharp corners=southeast}{#4}}
}
\newcommand{\rightbubble}[3]{%
  {\raggedleft
  \bubble{#1}{#2}{#3}{sharp corners=southeast}}
}
\newcommand{\leftbubble}[3]{%
  \bubble{#1}{#2}{#3}{sharp corners=southwest}
}
    
\usepackage{xspace}

\newcommand{\challenge}[1]{(\textbf{Challenge #1})\xspace}

\newcommand{\tabtext}[1]{
  \begin{varwidth}{.25\textwidth} #1  \end{varwidth}
  \vspace{4pt}
}
\newcommand{\tabtextvar}[2]{
  \begin{varwidth}{#2\textwidth} #1 \end{varwidth}
  
}

    \title{From Words to Worlds: Transforming One-line Prompt into Immersive Multi-modal Digital Stories with Communicative LLM Agent}
    \author[1,*]{Samuel S. Sohn}
    \author[1,*]{Danrui Li}
    \author[1,*]{Sen Zhang}
    \author[1]{Che-Jui Chang}
    \author[1]{Mubbasir Kapadia}
    \affil[1]{\small Rutgers University, 08854, New Jersey, USA}
    \affil[*]{Equal contribution}

    \date{}

\begin{document}

\markboth{Samuel S. Sohn, Danrui Li, Sen Zhang et al.}{From Words to Worlds}

\maketitle
\begin{figure}[htb]
  \centering
 \includegraphics[width=\linewidth]{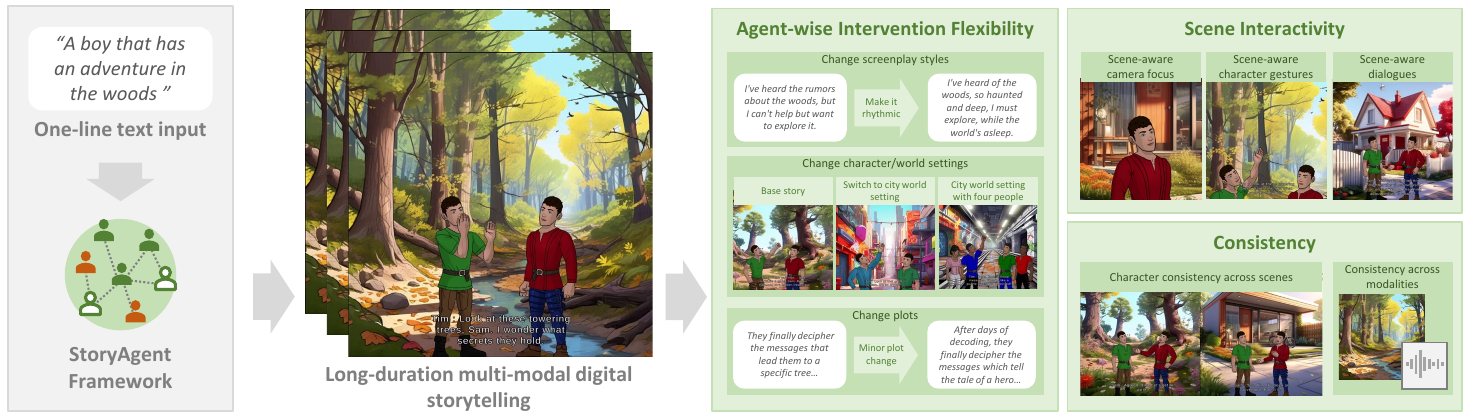}
  \caption{\textbf{StoryAgent} is a digital storytelling generation framework that integrates communicative Large Language Model agents with state-of-the-art generative models and tools. Taking one-line text instruction as input, it produces digital storytelling content with scene interactivity, long-duration consistency, and intervention flexibility.}
\label{fig:teaser}
\end{figure}

\begin{abstract}
Digital storytelling, essential in entertainment, education, and marketing, faces challenges in production scalability and flexibility. The StoryAgent framework, introduced in this paper, utilizes Large Language Models and generative tools to automate and refine digital storytelling. Employing a top-down story drafting and bottom-up asset generation approach, StoryAgent tackles key issues such as manual intervention, interactive scene orchestration, and narrative consistency. This framework enables efficient production of interactive and consistent narratives across multiple modalities, democratizing content creation and enhancing engagement. Our results demonstrate the framework's capability to produce coherent digital stories without reference videos, marking a significant advancement in automated digital storytelling. 

    \textbf{Key Words: }Pedestrian Model, Visual Attention, Retail Environment, Transportation hubs
\end{abstract}

\section{Introduction}

Digital storytelling has emerged as a powerful medium across various domains, including entertainment, education, and marketing \cite{de2017digital, lambert2013digital, WU2020103786} due to its ability to combine multimedia elements such as text, images, audio, and video to create immersive and interactive digital narratives.
Its versatile applications, ranging from interactive narratives in video games to immersive experiences in virtual reality, make it an indispensable tool for conveying information and eliciting emotions in today's digital age. Traditionally the production processes of digital storytelling narratives are often time-consuming and resource-intensive, limiting the speed and volume of content output.

Hence, there arises a critical need to streamline and automate the production pipeline to meet the growing demands for engaging and dynamic storytelling content. Recent advancements in text-based generative models\cite{dalle3, li2024playground, esser2024scaling, ho2022imagen, liu2023wavjourney, videoworldsimulators2024, kreuk2023audiogen}, which enable synthesizing assets in text, image, sound, and motion, have facilitated a hands-free digital storytelling production. It can democratize the creation process and potentially enable anyone who lacks the artistic skills to produce complex digital narratives easily. Several previous works \cite{cavazza2007madame, canvas2016, louarn_cinematography, Marti2018CARDINAL} tried to consolidate an automation procedure for part of the production pipeline for different use cases such as immersive interactive storytelling, storyboard making, staging, and script writing and functioning as authoring tools. With the help of multimodal generative models, we could ease the procedure of producing assets for digital storytelling. However, significant challenges persist in leveraging those multimodal generative models. 

One notable challenge is the need for flexible intervention \challenge{1}, as human creators often require the ability to modify initial generation results according to their preferences. Text-based generative models\cite{ ho2022imagen, videoworldsimulators2024} are proficient at creating high-quality short clips but offer limited fine-grained control over outcomes, such as modifying characters while maintaining the same storyline. Conversely, procedural methods \cite{canvas2016, kapadia_event_story} enable fine-grained control but typically require a specialized interaction interface with the framework, which often lacks a universal and convenient approach for human intervention.
Moreover, orchestrating the interactions between characters, objects, and scenes remains a difficult task \challenge{2}, yet indispensable to improving the visual fidelity and elevating the digital storytelling experience \cite{yi2024generating}.
Finally, consistency is needed for audience engagement. For instance, character appearances and voice tones should remain consistent with the narrative context throughout the story.
In addition, consistency should also include the synchronization of the textual plots and downstream modalities, including audio, speech, and visuals \cite{storysurvey, chang2023importance, chang2022disentangling, chang2022ivi, Salselas2019sound, Cummings2016immersion}. Despite advancements in diffusion-based animation generation \cite{feng2023dreamoving, guo2023animatediff, liew2023magicedit}, existing methods struggle to ensure long-term consistency or require additional inputs like reference videos or skeletons \challenge{3}.

We propose a novel StoryAgent framework, which integrates communicative Large Language Model agents \cite{li2023camel, wu2023autogen} with generative models and tools.  Our framework operates by initially drafting the story by a top-down approach, using communicative LLM agents to decompose text instructions into a hierarchical textual representation of the digital storytelling content, where the leaf nodes are descriptions of a single modality for a snippet of the timeline. Subsequently, it employs generative models and tools in a bottom-up fashion to create and assemble the corresponding assets from text descriptions.

The framework addresses the three aforementioned challenges. First, its textual representation and generation pipeline facilitate fine-grained control and intervention for human developers through simple natural language instructions \challenge{1}. Leveraging the reasoning capabilities of Large Language Models, the framework can comprehend instructions to identify and modify the corresponding leaf nodes in the hierarchy, while keeping other components unchanged. This process allows for targeted adjustments without disrupting the rest of the content.
Moreover, by combining the bottom-up idea in the procedural generation pipeline with the top-down hierarchical textual representation, our framework handles the issue of consistency and scene interactivity \challenge{2}. For instance, a character’s appearance across time frames is linked to the same costume asset ID within the textual representation. During video rendering, this asset is consistently reused, ensuring visual uniformity across all scenes. Similarly, semantic and spatial information from the generated images is captured and integrated into the textual hierarchy, paving the way to the scene interactivity for downstream components.
Inherently, with the design of our framework, it can generate videos that require no reference videos as inputs, and the temporal limitation does not exist\challenge{3}. 

Additionally, by leveraging text as the intermediate product to decouple story drafting and asset generation, our framework facilitates a plug-and-play structure. It not only allows for unprecedented coverage of modalities (see Tab. \ref{tab:mod}), but also easy integration of the latest generative models, which ensures that our framework performance can continuously benefit from ongoing research developments.

\begin{table*}
\centering
\begin{tabular}{cccccccccc}
\toprule
                        &         &\multicolumn{2}{c}{World}&\multicolumn{2}{c}{Character}&\multicolumn{3}{c}{Audio}&Others\\
                        \cmidrule(r){3-4}\cmidrule(r){5-6}\cmidrule(r){7-9}\cmidrule(r){10-10}
                        & Plot    & Semantic  & Visual    &Appearance & Animation & Music    & Speech   & SFX   &Cinematography\\
\midrule
\cite{Ammanabrolu_2020} &\checkmark&          &           &           &            &          &          &&\\
\cite{narrativeworlds}  &          &\checkmark&           &           &            &          &          &&\\
\cite{Hartsook_world_setting}&     &\checkmark&$\triangle$&           &            &          &          &&\\
\cite{louarn_cinematography}&      &          &           &           &            &          &          &&\checkmark\\    
\cite{zhang_write_animation}&      &          &$\triangle$&$\triangle$&\checkmark+ $\triangle$&          &          &&\\
\cite{Kumaran_2023}     &\checkmark&          &$\triangle$&$\triangle$&$\triangle$ &          &          &          &\\
\cite{liu2023wavjourney}&          &          &           &           &            &\checkmark&\checkmark + $\triangle$&\checkmark&\\
Ours                    &\checkmark&\checkmark&\checkmark &\checkmark + $\triangle$ &$\triangle$ &\checkmark&\checkmark&\checkmark + $\triangle$& \checkmark\\
\bottomrule
\end{tabular}

\caption{Digital storytelling components covered in prior works and ours. \checkmark = generative models. $\triangle$ = retrieval methods.}\label{tab:mod}
\end{table*}

\section{Background}

\subsection{Digital storytelling as authoring tools}

Digital storytelling has traditionally focused on providing authoring tools for human developers, typically adopting a procedural generation approach rather than end-to-end solutions like \cite{videoworldsimulators2024}. This methodology incorporates a wide range of storytelling components. Following the definition of \cite{storysurvey}, the components can be categorized into plots (textual contents such as story arc and events) and space (world settings, characters, scene props, etc.).

Previous storytelling authoring tools like \cite{cavazza2007madame, canvas2016, louarn_cinematography, Marti2018CARDINAL} are made for a specific stage in the film production pipeline, to involve human invention and refinement, they require domain knowledge for users.  Our StoryAgent framework provides agent-wise human intervention both on the high and low levels for amateurs and professionals. 

\subsection{Enable scene interactivity}

Recent studies have advanced the dynamics of character interactions within 3D environments, significantly aided by the explicit spatial representations of 3D objects \cite{zhang_write_animation, yi2024generating, chang2024learning, chang2024equivalency, chang2023importance, AdaptNet, Neural_character-scene_interactions, Physical_Character-Scene_Interactions}. In contrast, for 2D art styles, despite diffusion-based models delivering superior visual quality, the lack of inherent spatial scene information within 2D images poses challenges for implementing interactivity.

Nevertheless, existing research in image understanding offers many tools for deducing spatial data from images, including techniques like segmentation \cite{xie2021segformer} and depth estimation \cite{bhat2023zoedepth}. These methodologies can underpin a procedural pipeline designed to facilitate interactivity in 2D contexts. This paper represents an initial effort to enable such scene interactivity for 2D art styles.

\subsection{Consistency in digital storytelling}

Prior works have explored various approaches to ensure coherence and consistency within the plot generation, such as event-driven \cite{kapadia_event_story}, persona-driven \cite{zhang2022personaguided, xu_psychology_story},  state-space planning\cite{town_story}, and top-down decomposition \cite{Kim_multimodalstory}.
However, for the consistent joint generation between plot and space components, previous studies have been limited, generally focusing on one or two areas such as crowd motion \cite{chen_groups_from_language}, cinematography \cite{louarn_cinematography}, world settings \cite{Hartsook_world_setting,merino_five_dollar} etc. (see Tab. \ref{tab:mod}).
While some approaches have aimed to address the alignment of all visual components simultaneously, typically using latent diffusion text-to-image models \cite{maharana2022storydalle, shen2023storygptv, liu2024intelligent}, their effectiveness is still constrained by the time length, rendering them unsuitable for long-term digital storytelling scenarios. In this work, we aim to achieve a wide range of consistency between various components from a procedural approach.

\subsection{Generative digital storytelling with LLM agents}

Prior works have explored LLM-assisted generation frameworks covering several components such as text\cite{Ammanabrolu_2020, narrativeworlds}, audio\cite{liu2023wavjourney}, and visual\cite{Kumaran_2023}, where LLMs have shown their capabilities in assisting single-modal asset generation in the following aspects: 

Firstly, they can extract information from natural language descriptions and convert it into formatted parameters \cite{qing2023storytomotion, Kumaran_2023}.
Leveraging their world knowledge, LLMs can also break down complicated concepts into several simpler components \cite{liu2023wavjourney}, which lowers the difficulties for downstream generative models. In addition, the organization of components can be stored and updated in explicit formats \cite{delatorre2024llmr}. Finally, with reasoning capabilities \cite{yao2023react, wei2023chainofthought}, LLMs can plan for the generation tasks with predefined toolsets and real-time feedback. But prompt-based narrative scene generation tools like \cite{Kumaran_2023} are limited in the ability to generate multiple coherent and compelling narrative scenes in sequence.

LLM agent systems are widely used in complex generation tasks such as game world narratives \cite{park2023generative} and computer programs \cite{qian2023communicative}.
In these cases, the consistencies are achieved by predefined hierarchical memories and reflection procedures. 
In this paper, we apply an LLM agent system to ensure consistency in both temporal and modal dimensions.

\section{Method}

\begin{figure*}
  \centering
  \mbox{} 
  \hfill
  \includegraphics[width=\linewidth]{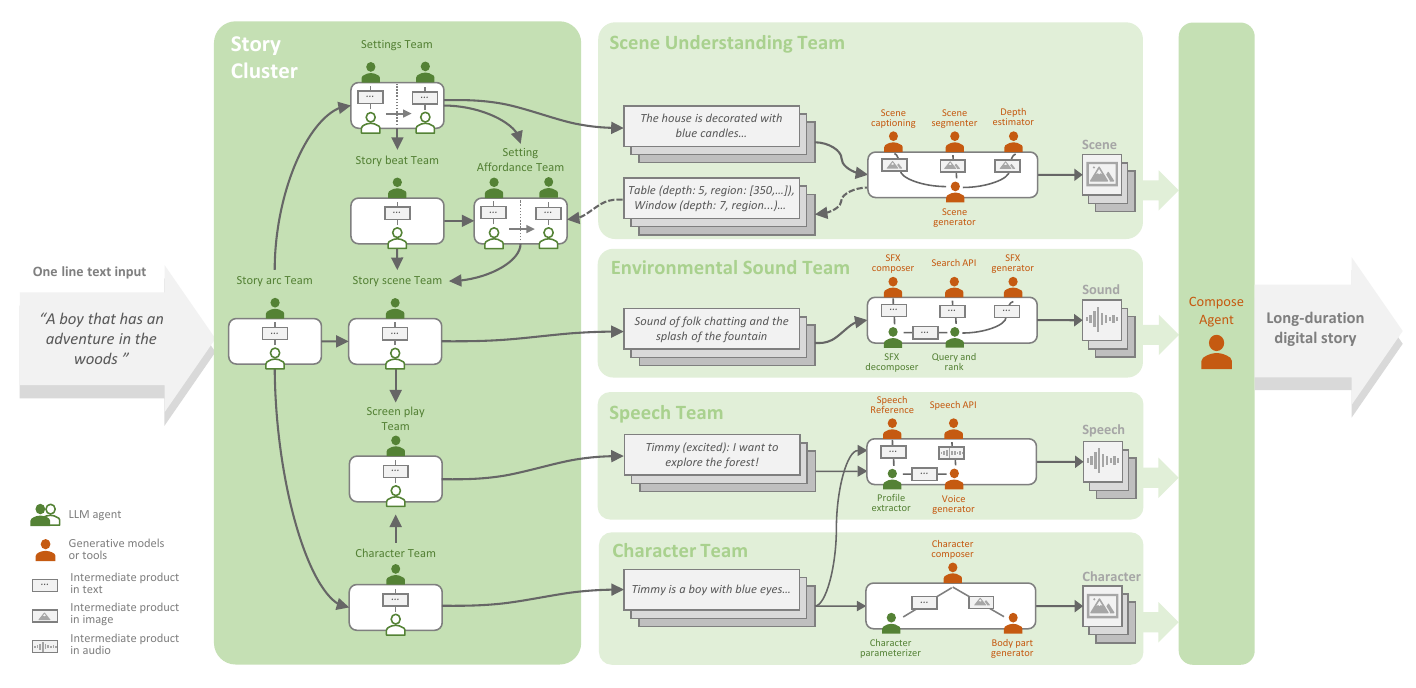}
  \hfill \mbox{}
  \vspace{-20pt}
  \caption{\label{fig:pipeline}%
           The framework of StoryAgent. Beginning with a text instruction, the framework builds the story with task decomposition, specifying all asset files for each modality in the textual description. Then generative models and tools are organized to create and compose tangible assets of the story. }
\end{figure*}

The entire pipeline (see Fig. \ref{fig:pipeline}) begins with a text instruction, then develops an intricate storyline by an LLM-agent-based story cluster (see \ref{sec:story-cluster}) with scene understanding capabilities (see \ref{sec:scene_understand}). The story cluster deconstructs the digital storytelling task into multiple subtasks, each targeting a specific modality. For each modality, the story cluster specifies all output asset files of the story through the textual descriptions. Based on these descriptions, generative models and tools are organized into asset generation teams (see \ref{sec:asset_cluster}) to create tangible assets of the story. Finally, approaches for potential interventions are introduced in \ref{sec:intervention-method}.

\subsection{Story cluster}
\label{sec:story-cluster}

The story cluster drafts the storyline through a network of LLM agent teams. Inspired by the pipeline workflow in the film industry, the agent teams handle story arcs, characters, settings, story beats, setting affordances, story scenes, and the screenplay respectively.

The network begins with a story arc team, which composes a blueprint of the narrative in text, instructing the entire production at the highest level.
Then, the blueprint is distributed to numerous downstream teams to handle different narrative components such as character settings, settings, scenes, screenplays, and so on. The intermediate products that are circulated between teams are always in text. Although the cluster is built upon AutoGen\cite{wu2023autogen}, the circulation follows a predefined procedure (see Fig. \ref{fig:teaser}) to ensure generation stability. 

\begin{figure}[htb]
  \centering
 \mbox{}
      \includegraphics[width=.8\linewidth]{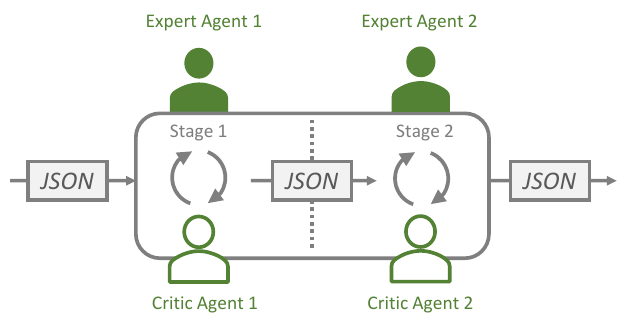}
  \mbox{}
  \caption{\label{fig:expert-critic} Structure of a two-stage LLM agent team. It takes upstream JSON as inputs and uses two expert-critic LLM agent pairs to process. Finally, another JSON string will be generated for downstream teams.}
  \vspace{-10pt}
\end{figure}

Based on upstream content, each team's generated outcome is incrementally optimized through single-stage or multi-stage dialogues (see Fig.\ref{fig:expert-critic}). All multi-stage configurations are manually designed to further decompose the generation process, as some agent teams require complex reflections that the current LLM backbone cannot handle. In each stage, the dialogue takes place between two LLM agents. Inspired by the role-playing method in \cite{li2023camel}, each agent is initialized with a system prompt, where dialogue history and upstream content are attached to an agent-specific template. 

Among the two, one ``expert agent'' is responsible for creating component specifications based on predefined requirements and upstream inputs. The other ``critic agent'', evaluates and scores them against predefined criteria. 
The roles of the agents, their constraints, and the output formats are shared between the expert and the critic within individual team stages but differ across different agent teams and stages. For comprehensive prompts for all agents, please refer to the appendix.

Starting from the expert agent, the two respond to each others' outputs in a round-robin way, where the circulation ends when the critic is satisfied with the expert's outcome.

Utilizing GPT-4 as the LLM agent backbone, this expert-critic design facilitates a robust capacity for self-correction, which ensures consistency between the upstream commands and downstream outcomes. To illustrate, consider the following example where the critic agent helps the expert agent refine a character profile:

\leftbubble{Admin}{bubblegray}{(provide a story arc in JSON format)}

\rightbubble{Expert in Character}{bubblegreen}{\{"name": "Village elders"...\}}

\leftbubble{Critic in Character}{bubblegray}{...The character name "Village elders" is plural and should be singular...The expert should break down "Village elders" into individual characters with singular names. For example, they could create characters for a few prominent elders and villagers.}

\rightbubble{Expert in Character}{bubblegreen}{\{"name": "Elder John"...\},\{"name": "Elder Mary"...\}...}

\leftbubble{Critic in Character}{bubblegray}{(pass the check)}

\noindent

\subsection{Image-based scene understanding and interaction}
\label{sec:scene_understand}

Simply putting the characters in the foreground and the story setting images in the background could result in unreasonable arrangements like a character standing on the water or up in the air.
To enhance the fidelity of each scene and give audiences a better immersion in the story, the pipeline conducts several measures: first fusing image semantic segmentation and depth estimation explicitly and implicitly as the visuospatial information and then letting setting affordance agents to enable foreground characters to interact with background images, which would otherwise be disjoint. The visuospatial information could also benefit cinematography by providing the object's estimated position and distance to the camera.

\subsubsection{Scene understanding}
The scene understanding process is structured in three key stages, designed to equip setting agents with comprehensive knowledge of the narrative environment, enabling them to provide relevant affordances for dynamic story interaction.

Initially, story setting agents construct a hierarchical graph of all narrative settings. Each node within this graph represents a distinct setting and includes essential details like the setting's name, visual prompts, and its relationship with other settings (parent and children). From these visual prompts, background images are generated using \cite{li2024playground}, forming the visual foundation of the story's environments.

The generated images undergo semantic segmentation to identify and categorize objects within each setting. This segmentation (\cite{xie2021segformer}), paired with depth estimation (\cite{zoedepth}), provides a three-dimensional understanding of each scene. Objects are encapsulated within bounding boxes, which highlight their spatial position and median depth, aiding in the precise placement within the narrative space. This crucial spatial data informs how characters and cameras will interact with these objects, ensuring accurate and realistic scene compositions(see \ref{sec:char_and_cam_interaction}).

\begin{figure}[htb]
  \centering
 \mbox{}
      \vfill 
      \includegraphics[width=1.0\linewidth]{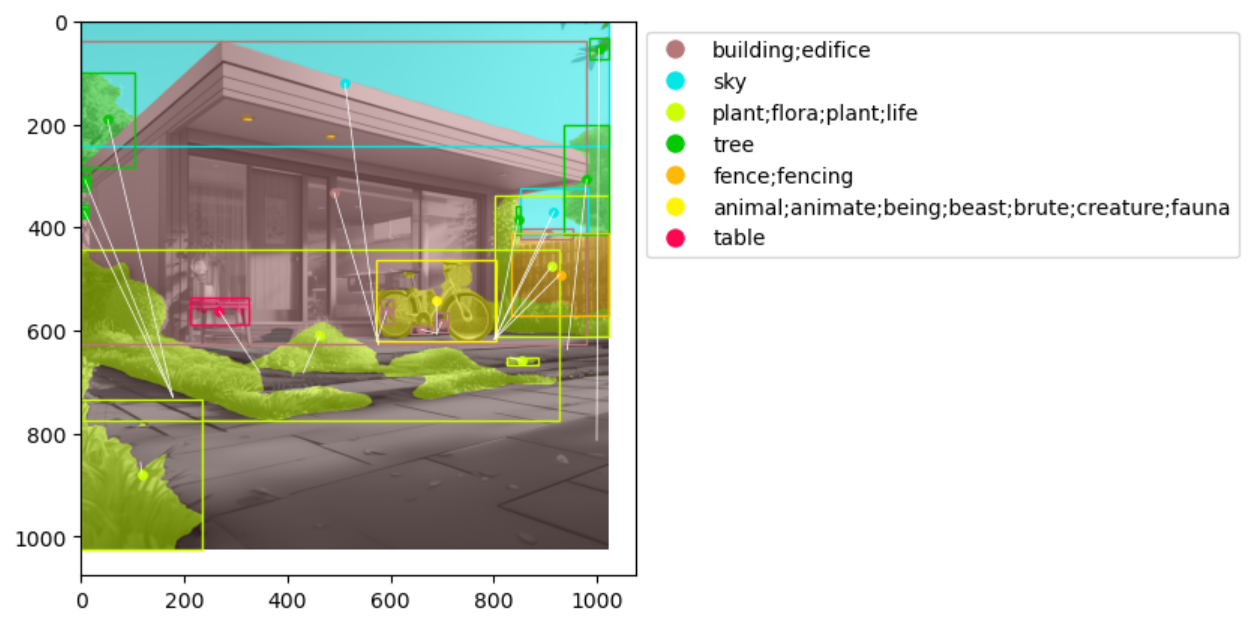}
      \vfill 
  \mbox{}
  \vspace{-10pt}
  \caption{\label{fig:scene_understand} Semantic scene understanding example}
\end{figure}

In the final stage, setting affordance agents utilize the detailed object and spatial data (see appendix \ref{app:affordance_output}) to create a rich layer of interaction possibilities through affordances. These affordances are meticulously documented with object relations, narrative relevance, and evidence of the object's existence in the image. This structural information is then passed to story scene agents, who use it to script interactions and narrative events, ensuring that characters can interact naturally with their environment.

\subsubsection{Character and camera interaction}
\label{sec:char_and_cam_interaction}

Our pipeline processes and utilizes object location information to enhance storytelling through precise cinematography. Once this data is captured, screenplay agents integrate it to orchestrate scene dynamics effectively. This involves using the object centers as focal points in animation and cinematography—characters interact with key objects via targeted movements, and cameras adjust focus and framing based on the object's position and estimated distance. This approach allows for the strategic selection of shot types (close, medium, or wide) to best capture the narrative moment.

\subsection{Asset generation teams}
\label{sec:asset_cluster}

Each agent generation team represents a hybrid of LLM agents and generative models. Generally, the LLM agents here serve as the bridge between the upstream textual descriptions and the downstream generative models and tools. 
Specifically, LLM agents aim to convert specifications articulated in natural language into model parameters and reflect on feedback from generative models and tools. 

\begin{figure}[htb]
  \centering
 \mbox{}
     \hfill
      \includegraphics[width=.3\linewidth]{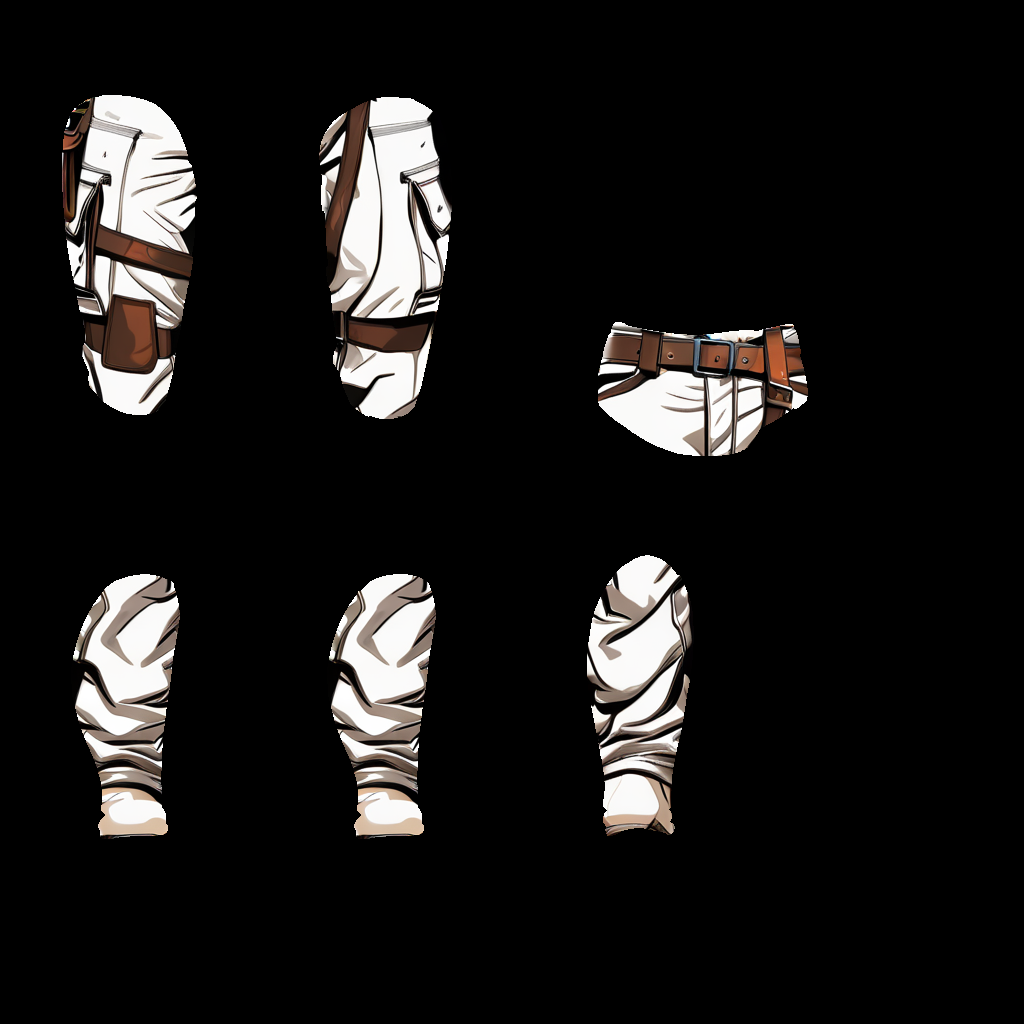}
      \hfill 
      \includegraphics[width=.3\linewidth]{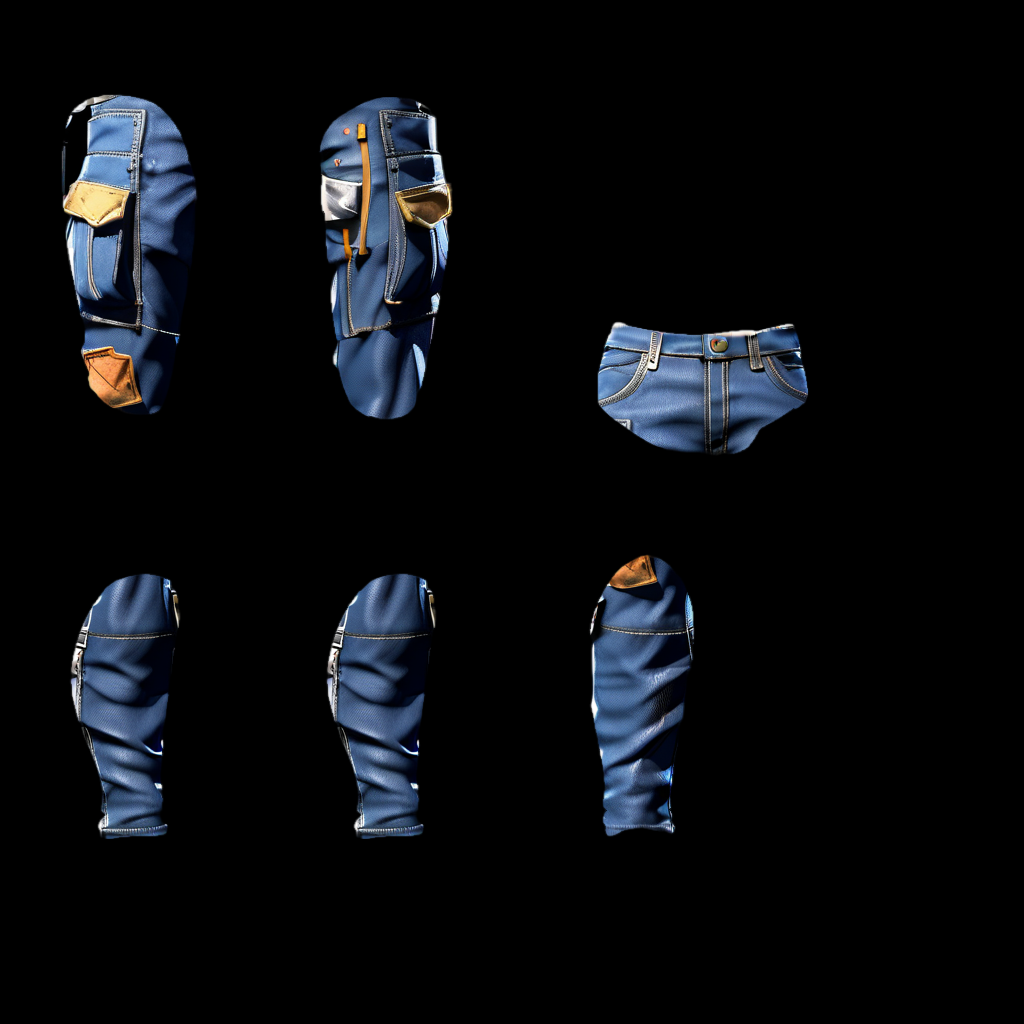}
      \hfill 
      \includegraphics[width=.3\linewidth]{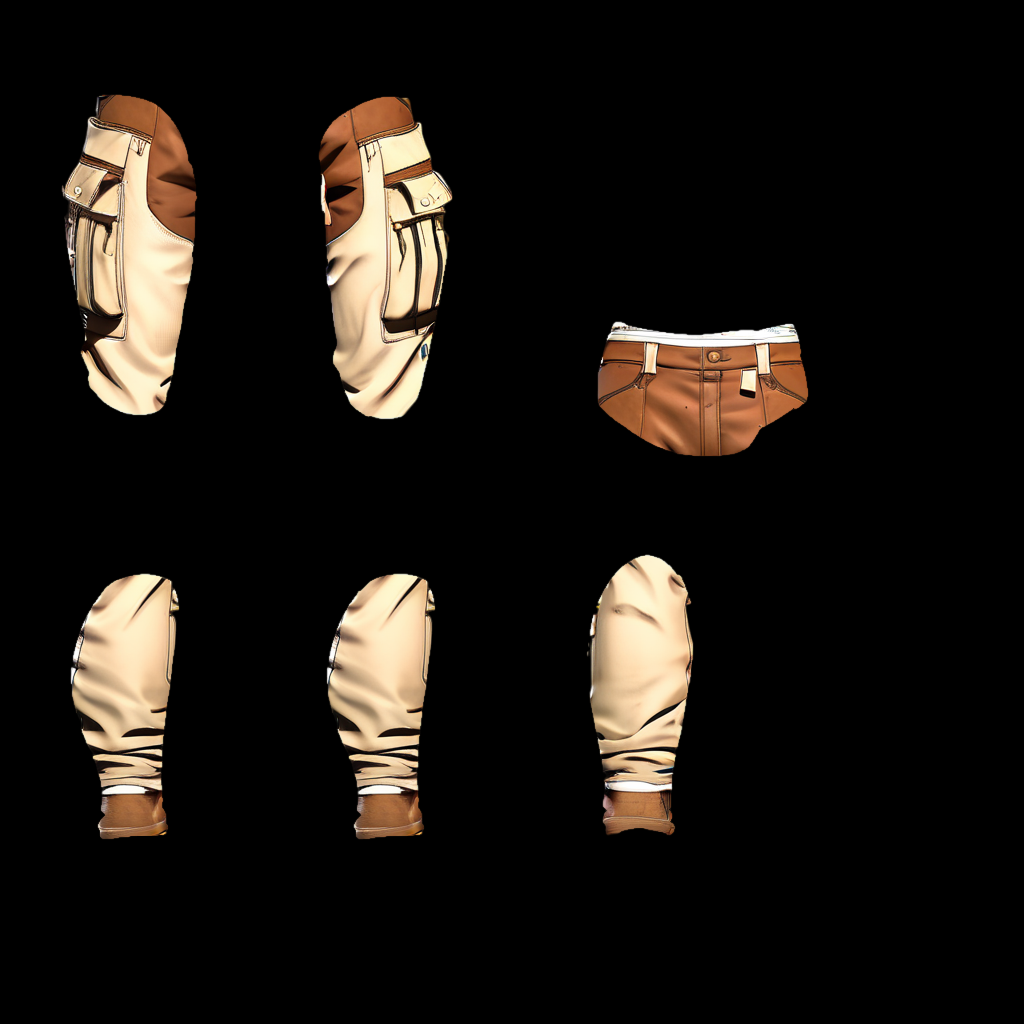}
      \hfill 
  \mbox{}
  \caption{\label{fig:ex3} Character pants asset examples}
\end{figure}

\begin{figure}[htb]
  \centering
 \mbox{}
     \hfill
      \includegraphics[width=.4\linewidth]{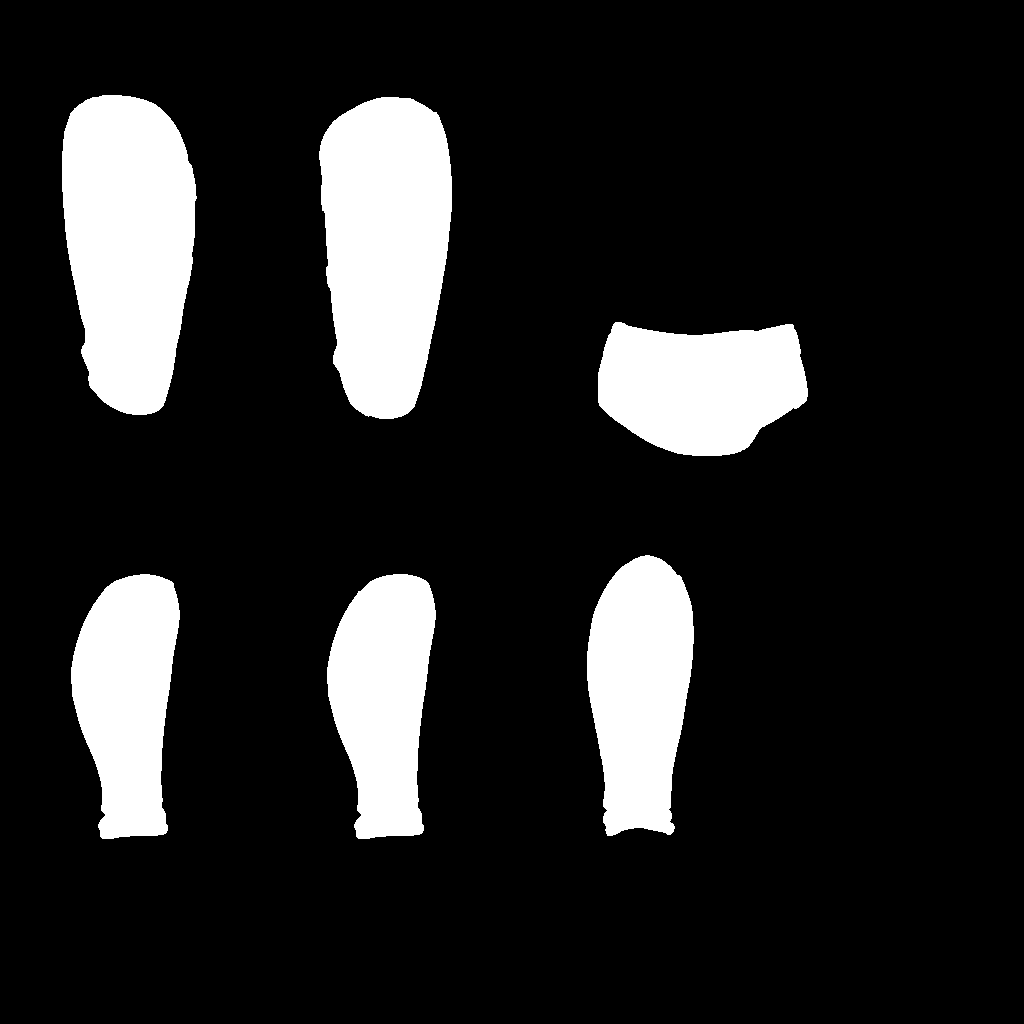}
      \hfill
      \includegraphics[width=.4\linewidth]{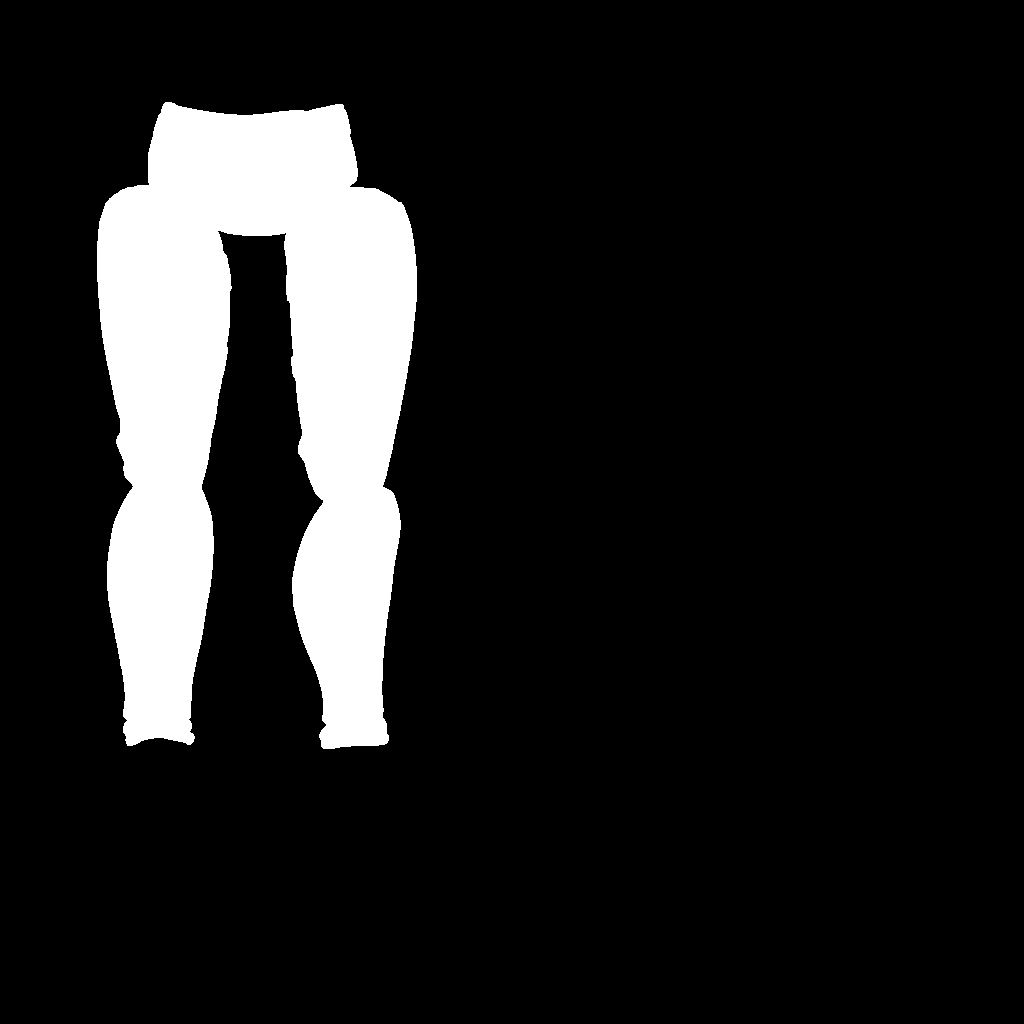}
      \hfill 
  \mbox{}
  \caption{\label{fig:mask} Character parts mask. Left is the original texture format and right is the assembled mask for better generation quality}
\end{figure}

\begin{figure}[htb]
  \centering
 \mbox{}
     \hfill
      \includegraphics[width=.6\linewidth]{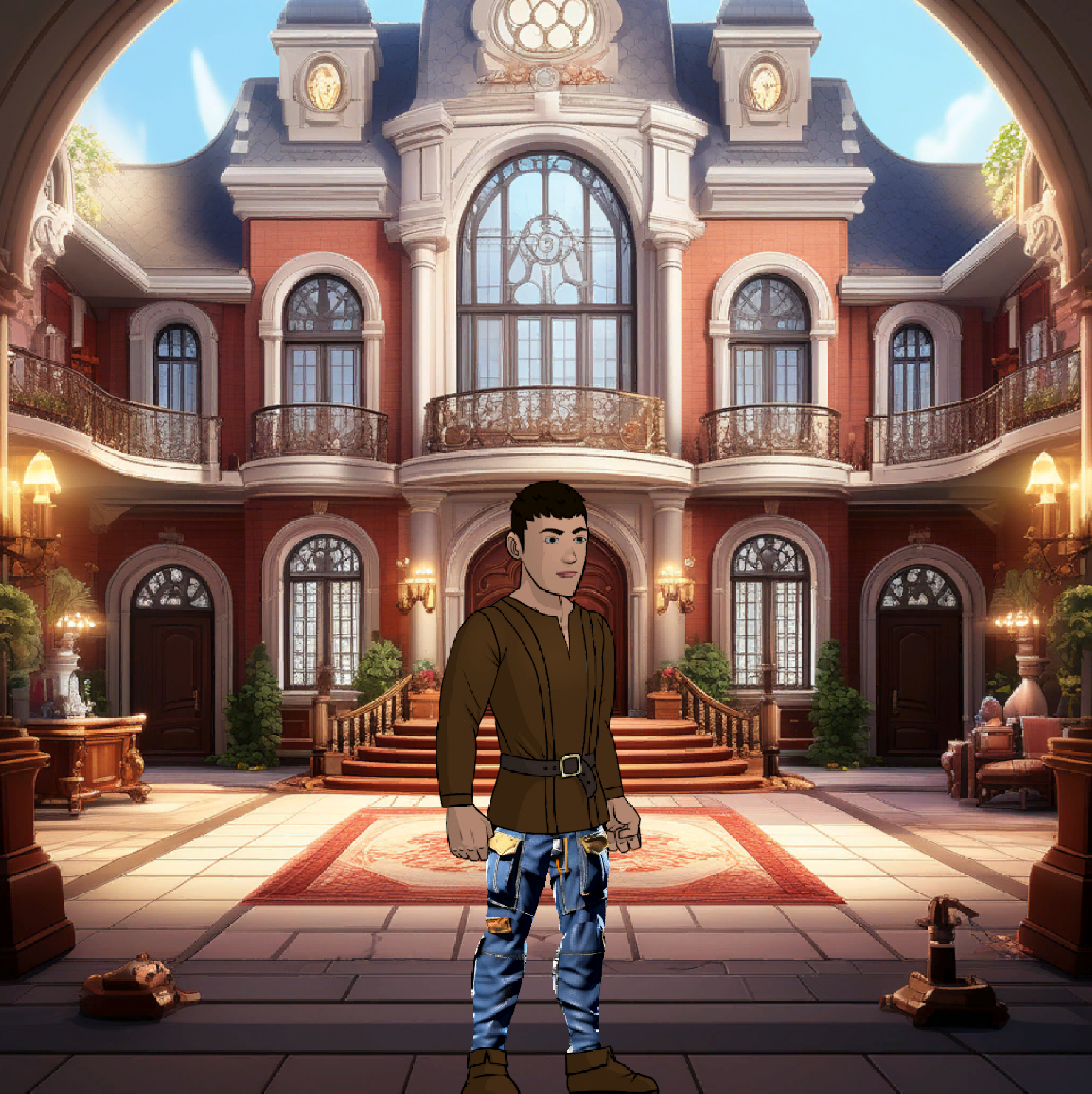}
      \hfill 
  \mbox{}
  \caption{\label{fig:pants} Pants asset in the runtime}
\end{figure}

In our study, we employ the rigging framework, CC2D\cite{cc2d}, to construct and animate human avatars. To generate image assets for different parts of the character, such as pants and clothing, we input appearance descriptions into a text-to-image generative model \cite{li2024playground}. These assets are produced using a supplemental prompt—such as ``detailed, cartoon, 8k''—to enhance the initial description. Since diffusion-based generative models are hard to generate meaningful 2D textures directly, the generative procedure goes through a composite and decompose process. First We composite the 2D texture of the tunic and pants into a meaningful shape of the asset (see Fig. \ref{fig:ex3}), then generate assets using its assembled mask. After the generation is done, we decompose the whole assets back with their texture mask and then integrate them within the CC2D framework (see Fig. \ref{fig:mask}). Textures are loaded during runtime (see Fig. \ref{fig:pants}).

For background image creation, we utilize the same generative model to ensure style consistency. We use \cite{compel} to lift the token limitation and generate backgrounds by visual prompts provided by the setting configuration agents. Furthermore, these images undergo further processing in a scene understanding cluster (see \ref{sec:scene_understand}) to enable interactivity. This additional step allows for dynamic interaction within the generated environments (detailed in \ref{sec:scene_understand}).

In speech generation, a character description simplified by LLMs is firstly fed to AudioGen\cite{kreuk2023audiogen} to create a voice fingerprint, containing the voice feature of that character. Then it is smoothed by XTTS \cite{coqui-aitts_nodate}, producing a reference speech audio. Next, the reference audio, speech in text, and the emotion state in text are fed to ElevenLabs API \cite{elevenlabselevenlabs-python_2024} to create the final speech audio. 

Inspired from \cite{liu2023wavjourney}, the sound description is firstly decomposed to components by LLMs, where different decomposition methods are tailored for sound effect types (looping or one-time). Then, for each component, an LLM agent is applied to search corresponding assets on FreeSound \cite{freesound}, where the assets are evaluated by feeding their descriptions to the agent. The search is repeatedly triggered with keywords that are optimized by the agent using previous search results. If there are no suitable assets, AudioGen\cite{kreuk2023audiogen} is called to create one.
All music clips are created by MusicGen \cite{kreuk2023audiogen}, a text-to-music generative model.

Consistency is optimized by communication between LLM agents and other generative models and tools. For instance, in sound retrieval, after the initial description is broken down into simpler components, each component's description is transformed into a query string, which is iteratively refined based on the results of previous searches. This method underscores the dynamic use of LLMs to enhance the consistency between the text description and the retrieved assets.

\leftbubble{Admin}{bubblegray}{The sound of townsfolk chatting, children playing, and the gentle splash of the fountain water.}

\rightbubble{Decomposition Agent}{bubblegreen}{
\{
    "name": "townsfolk chatting",
    "desc": "the sound of townsfolk chatting in a town square",
    "volume": -25
\},
\{
    "name": "children playing",
    "desc": "the sound of children playing in a town square",
    "volume": -20
\},
\{
    "name": "fountain water",
    "desc": "the gentle splash of fountain water in a town square",
    "volume": -30
\}
}
\rightbubbleat{Decomposition Agent}{bubblegreen}{the gentle splash of fountain water in a town square}{Query Agent}

\rightbubbleat{Query Agent}{bubblegreen}{water +fountain +splash +gentle -ocean -river}{Search API}

\leftbubble{Search API}{bubblegray}{(a list of sound with descriptions)}

\rightbubble{Query Agent}{bubblegreen}{water +fountain +gentle +town -ocean -river -greenway}

\leftbubble{Search API}{bubblegray}{(a list of sound with descriptions)}

\rightbubble{Query Agent}{bubblegreen}{fountain +gentle +town}

\leftbubble{Search API}{bubblegray}{(a list of sound with descriptions)}

\rightbubble{Query Agent}{bubblegreen}{(The index of the best sound asset)}

\subsection{Intervention}
\label{sec:intervention-method}

While our framework facilitates a hands-free generation process, it also offers flexible intervention approaches for human creators. The framework allows for the following intervention approach:

\begin{itemize}
    \item Full regeneration from the beginning: by feeding prior generation results and extra instructions to the framework inputs, the framework updates all contents with maximum consistency ensured. We apply this approach in Section \ref{sec:intervention}. Please refer to the instruction prompt in the appendix.
    \item Intermediate the process: by fixing upstream results and altering agent system prompt, the framework updates all downstream contents with finer control over selected modalities.
    \item Replacement: by replacing generated results with human-crafted ones, human creators get their maximum control over the outcome. However, replacing downstream assets may cause inconsistency issues since the framework cannot reflect on them.
\end{itemize}

\begin{figure*}[htb]
  \centering
 \mbox{}
     \hfill
     
      \begin{subfigure}[t]{0.18\textwidth}
        \includegraphics[width=\textwidth]{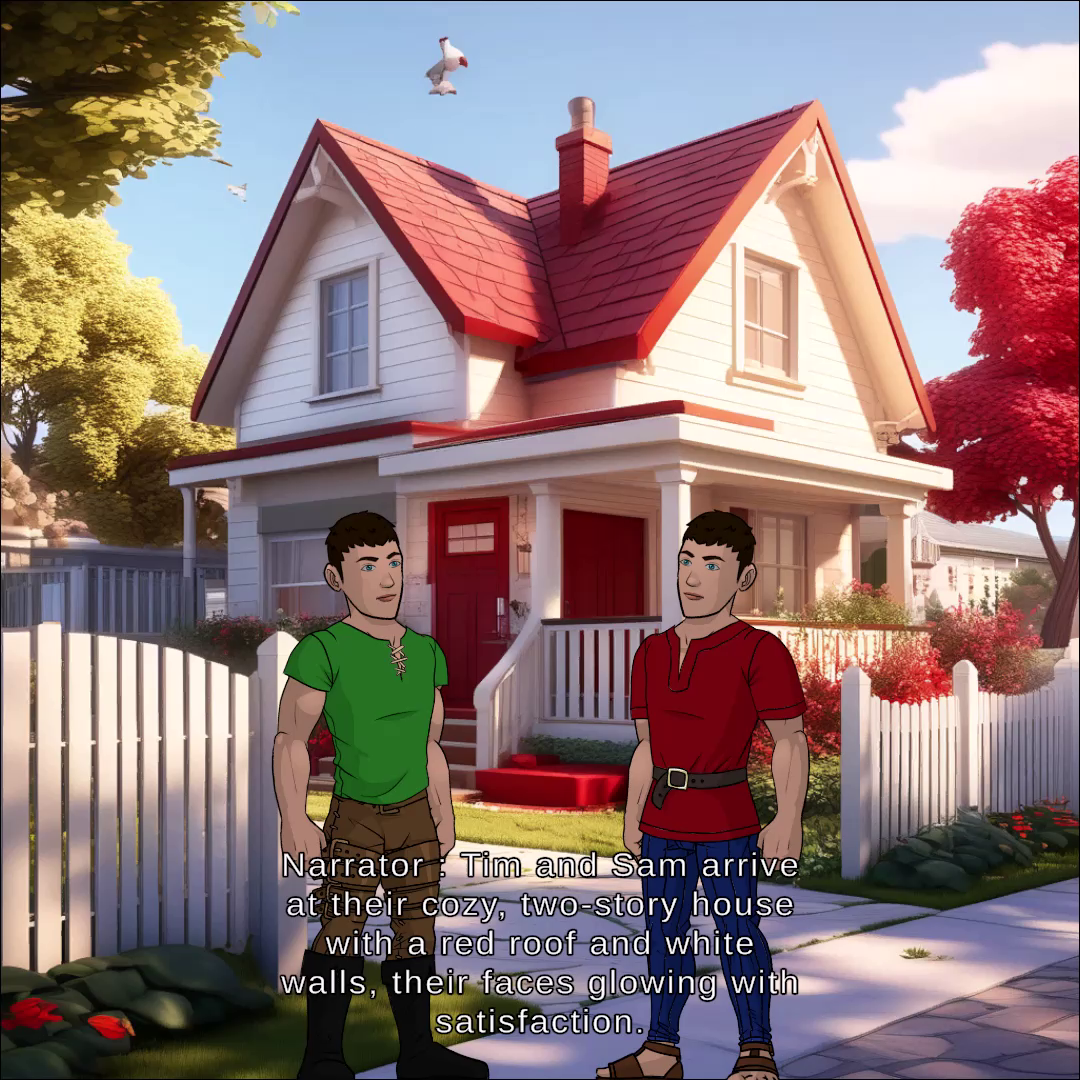}
         \caption{A curious boy named Tim and his adventurous friend, Sam, decide to explore the woods near their homes, despite the rumors of it being haunted.}
     \end{subfigure}
      \hfill
      \begin{subfigure}[t]{0.18\textwidth}
        \includegraphics[width=\textwidth]{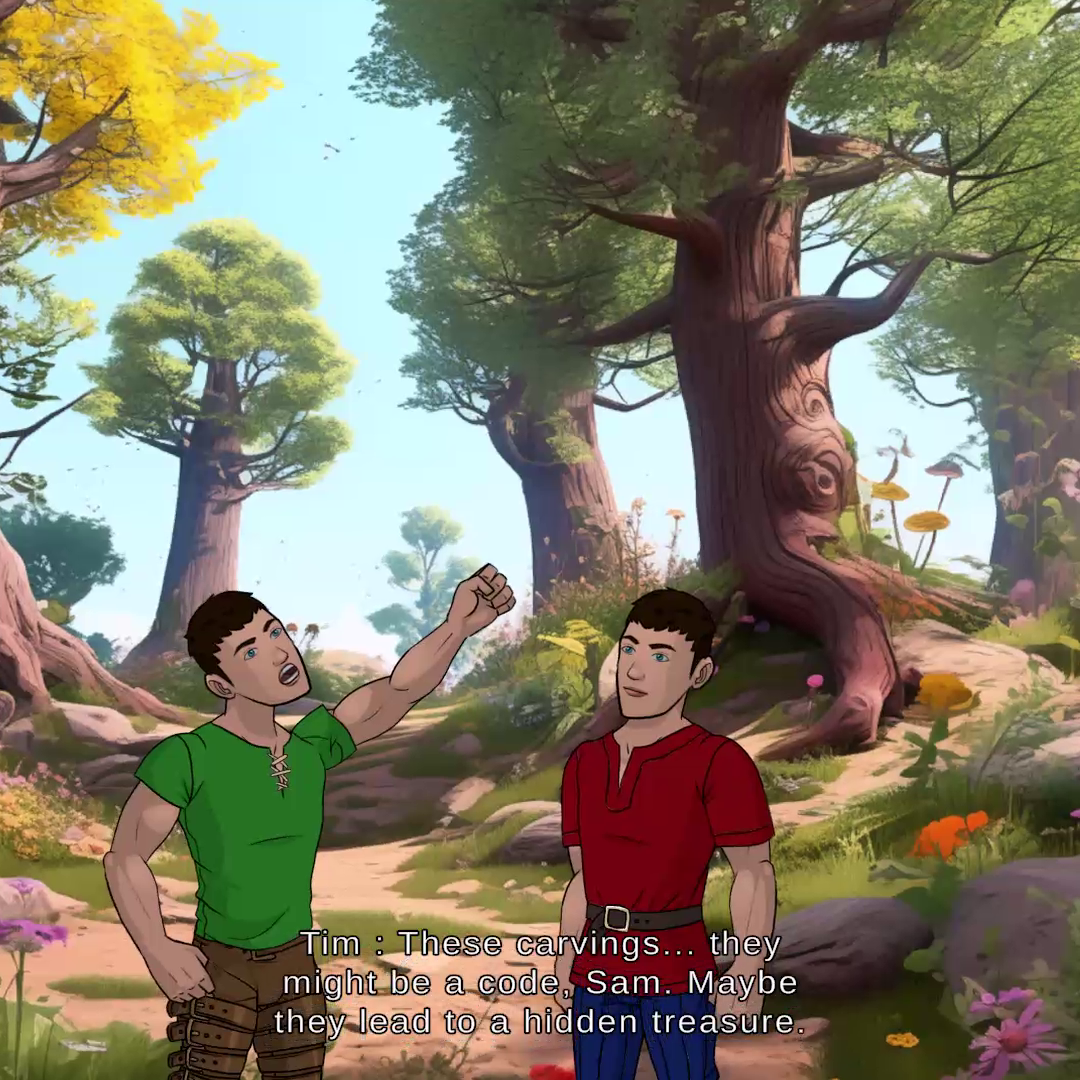}
         \caption{As they venture deeper into the woods, they find cryptic messages carved into the trees. They decide to decipher these, believing they might lead to a treasure.}
     \end{subfigure}
      \hfill
     \begin{subfigure}[t]{0.18\textwidth}
        \includegraphics[width=\textwidth]{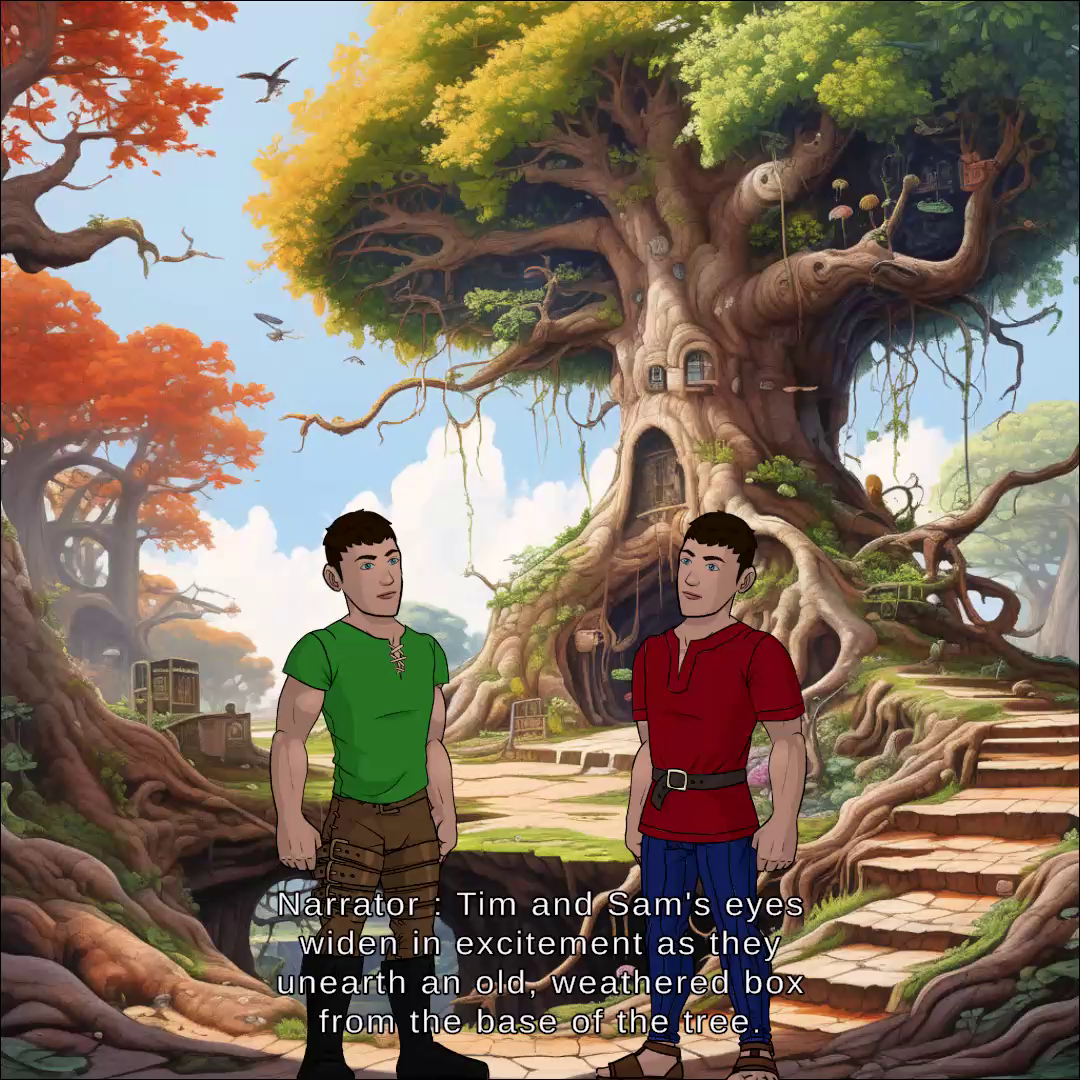}
         \caption{They finally decipher the messages that lead them to a specific tree. They find an old, weathered box buried at the base of the tree.}
     \end{subfigure}
      \hfill
    \begin{subfigure}[t]{0.18\textwidth}
        \includegraphics[width=\textwidth]{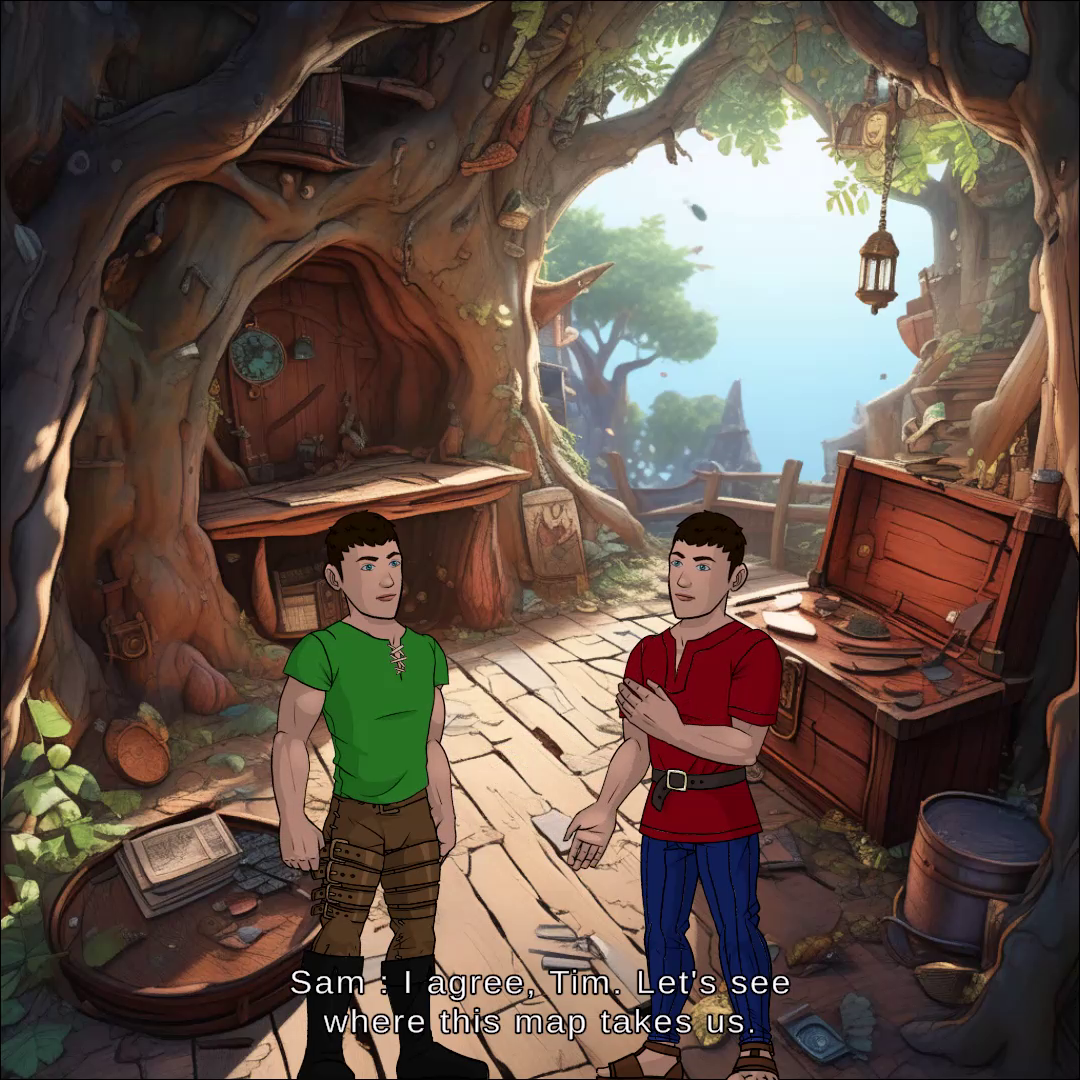}
         \caption{They open the box to find a map of their town from decades ago, with a path marked leading back to their homes.}
     \end{subfigure}
      \hfill
    \begin{subfigure}[t]{0.18\textwidth}
        \includegraphics[width=\textwidth]{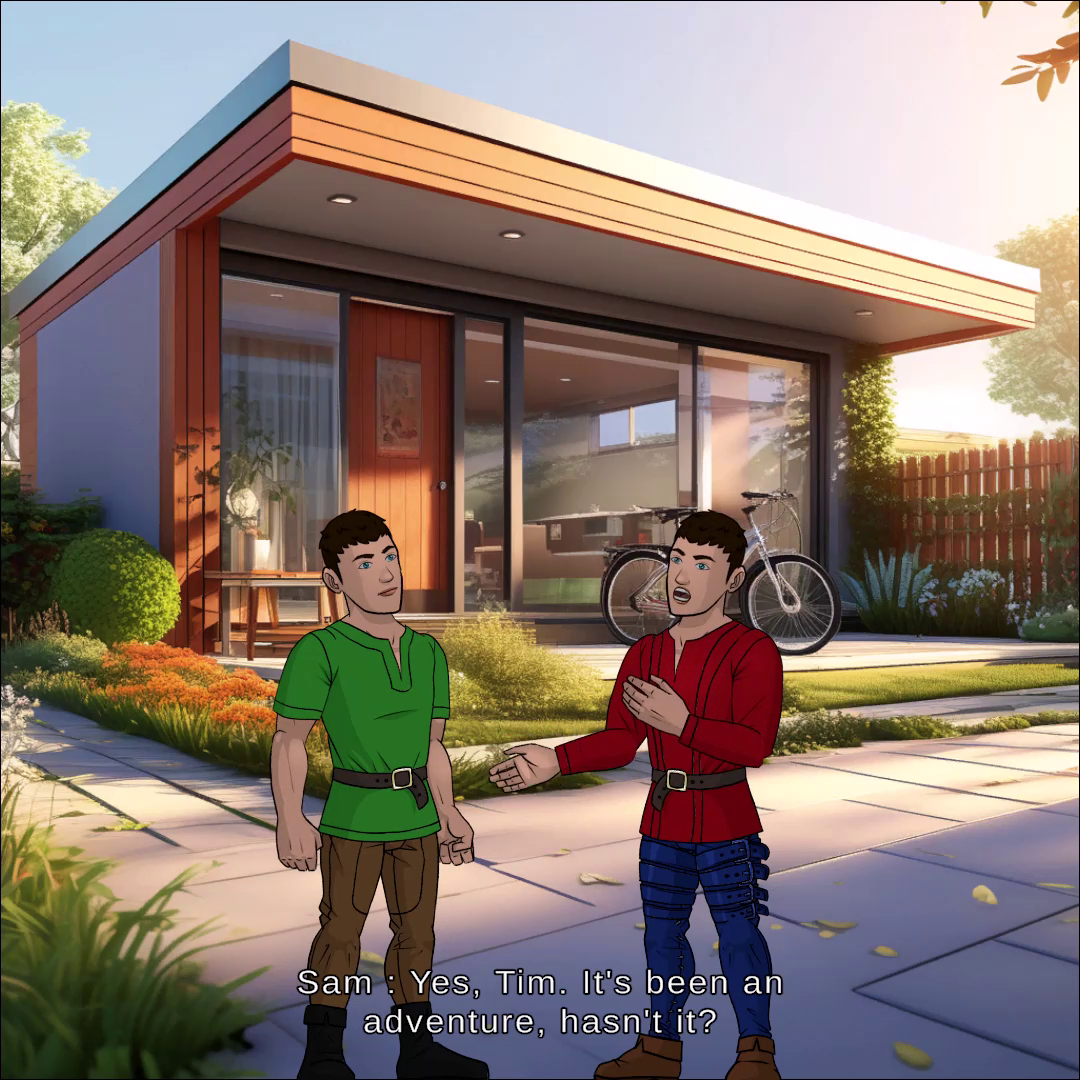}
         \caption{They follow the path on the map and realize that the treasure was the journey and the memories they made. They return to Sam's home with friendship strengthened.}
     \end{subfigure}
      \hfill
  \mbox{}
  \caption{\label{fig:base story} Screenshots of a story generation, with one frame selected from each stage of a five-stage storytelling arc.}
\end{figure*}

\section{Storytelling Experience}

In this section, we present the qualitative results of our pipeline, highlighting three benefits of our text-based pipeline.
We first demonstrate how the reasoning capabilities of StoryAgent facilitate consistency between plots and downstream components. Then we illustrate how the coordination of various agents and generative tools ensures scene interactivity in 2D art styles. Finally, we show its capability to adapt human intervention during the generation process, thus providing story alternatives for human creators. Here is the story arc for our baseline story: 

In the story, Tim and his friend Sam explore the woods near their homes, discovering cryptic messages on trees that they believe might lead to treasure. Following the clues, they uncover an old box containing an antique map of their town. The map guides them back home, revealing that the real treasure was the adventure and the bond they strengthened along the way.

\subsection{Story consistency}

\begin{figure}[htb]
  \centering
 \mbox{}
     \hfill
      \includegraphics[width=.3\linewidth]{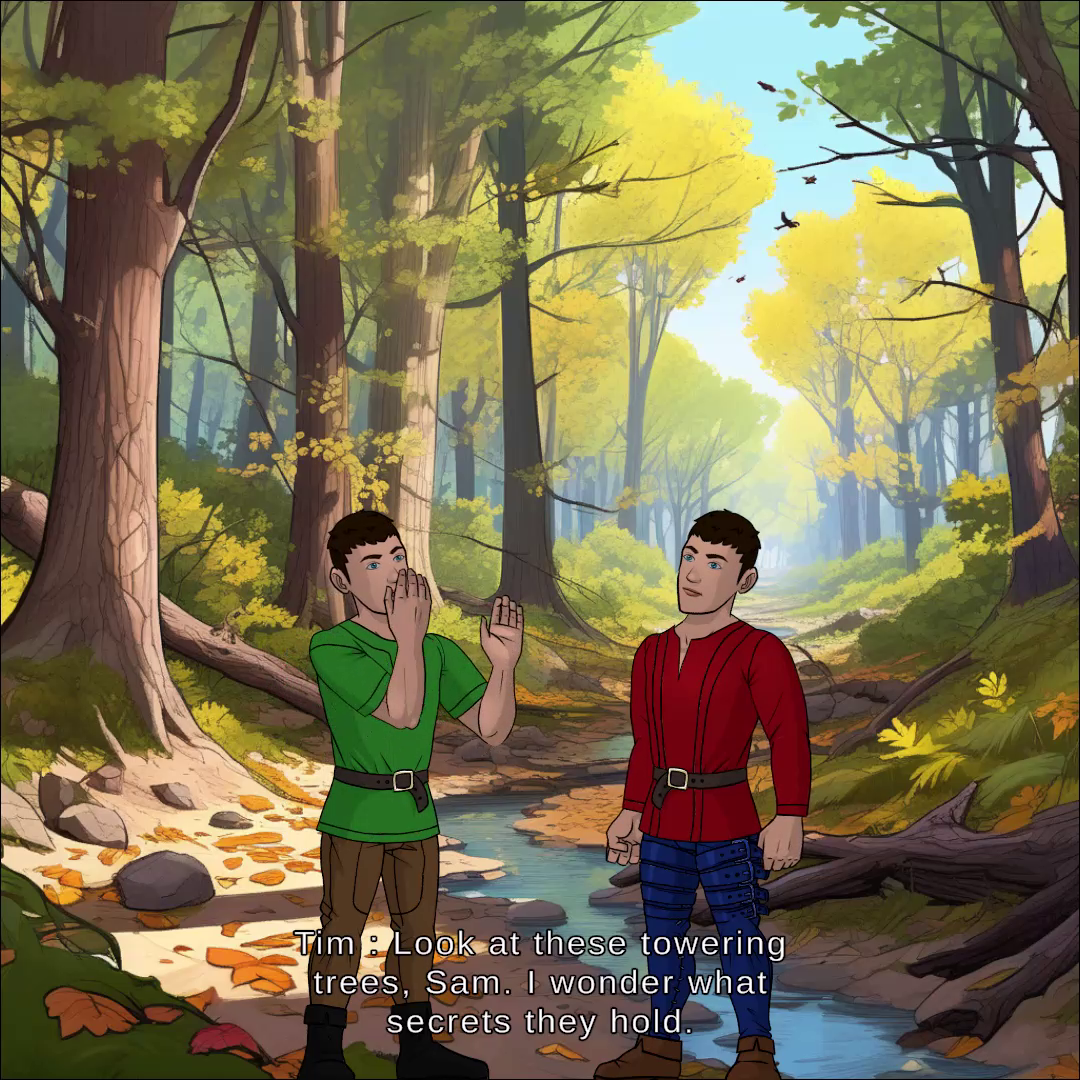}
      \hfill
      \includegraphics[width=.3\linewidth]{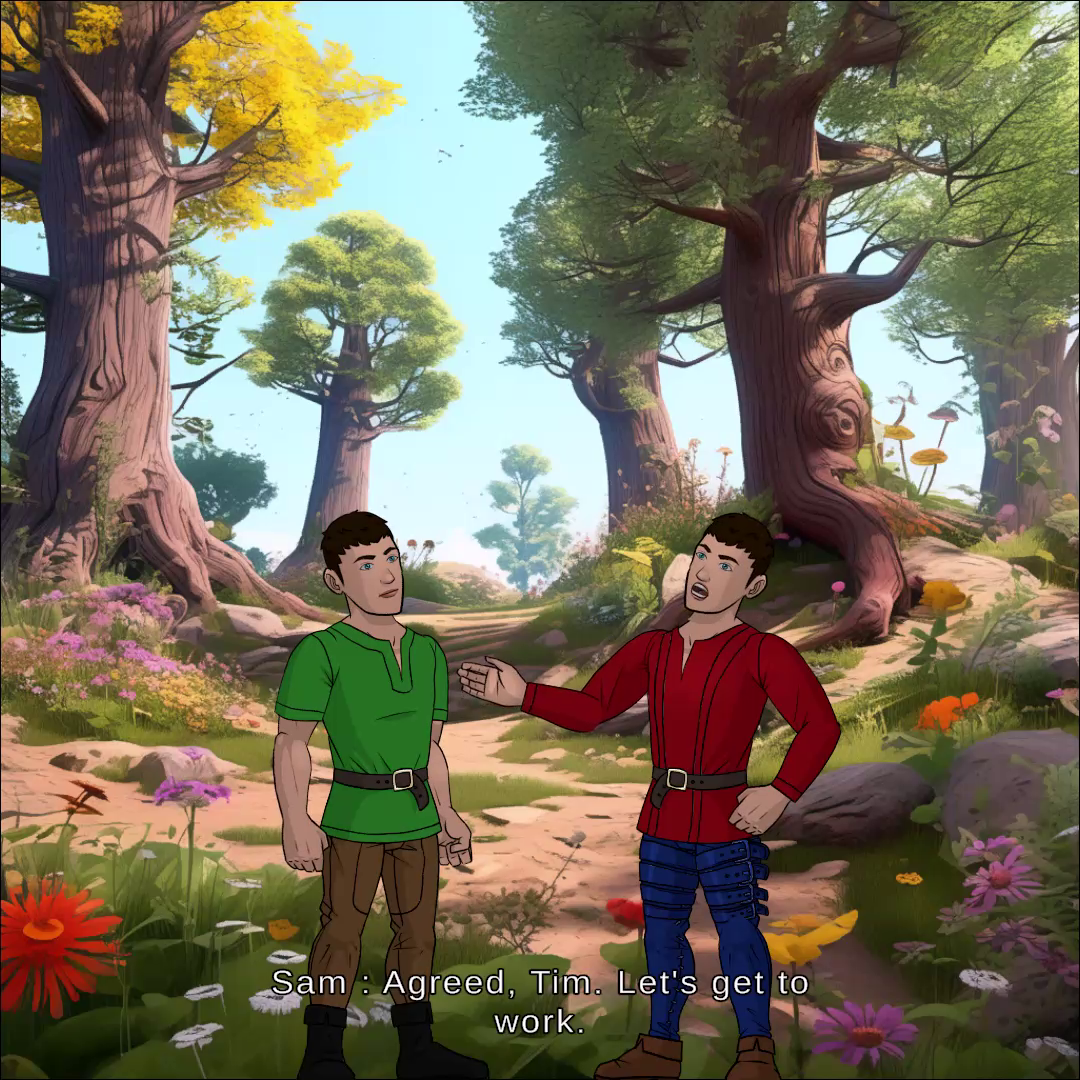}
      \hfill 
      \includegraphics[width=.3\linewidth]{figs/consistency__3_.png}
      \hfill 
  \mbox{}
  \caption{\label{fig:consistency} Character consistency across scenes. Based on the textual descriptions of the generative narrative, the character assets are reused across scenes, ensuring visual consistency.}
\end{figure}

Our framework keeps a hierarchical organization of all asset files, which enables asset re-usages. Consequently, this structure ensures appearance consistency across all visual elements, both within neighboring frames and across different scenes (see Fig.\ref{fig:consistency}).

\begin{figure}
  \centering
 \mbox{}
    \begin{minipage}[c]{0.4\linewidth}
       \includegraphics[width=\linewidth]{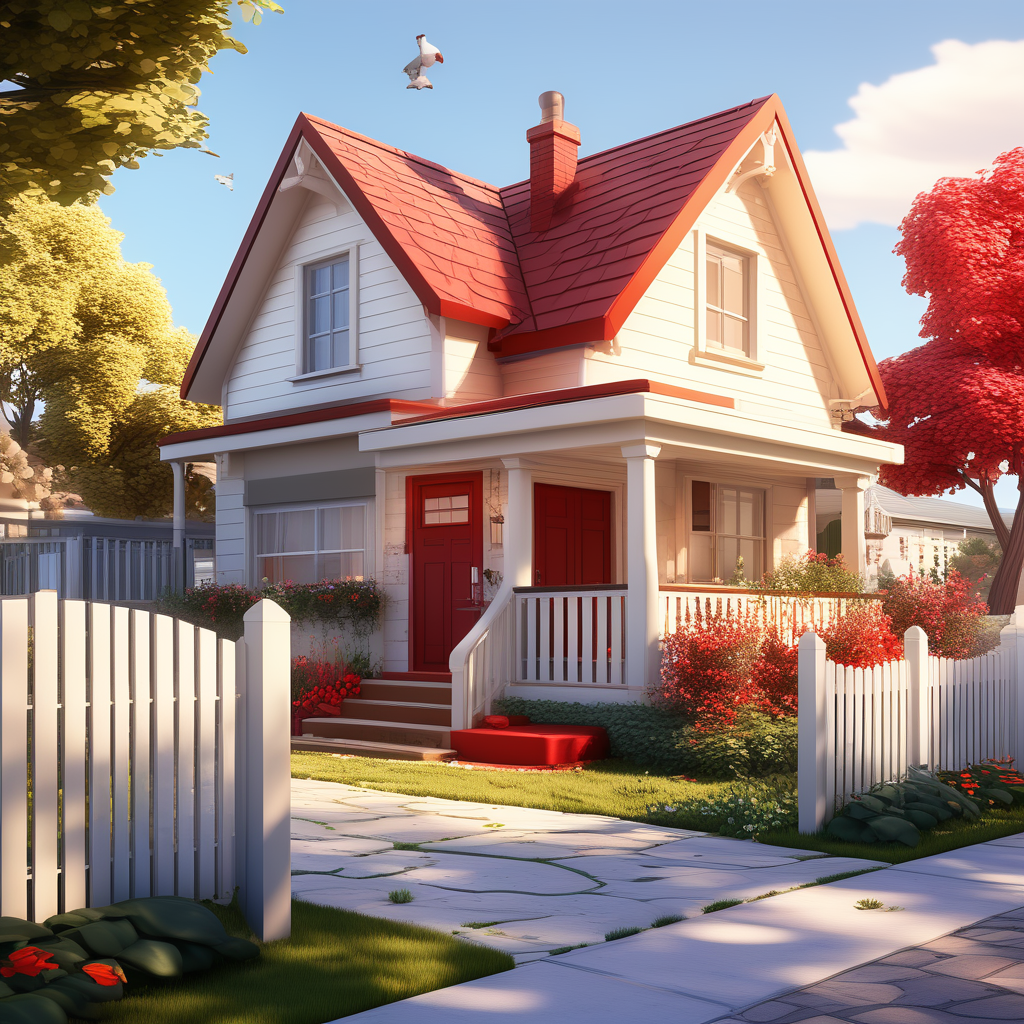} 
    \end{minipage}
    \hfill 
    \begin{minipage}[c]{0.56\linewidth}
        \tightscript{
[Visual] A small room with a single bed, a wooden desk, a bookshelf filled with books, a window overlooking the yard, and a blue rug.

[Audio] Background sounds of a quiet suburban neighborhood, with distant sounds of children playing and birds chirping.
        }{}
    \end{minipage}
    \hfill
    \begin{minipage}[c]{0.4\linewidth}
       \includegraphics[width=\linewidth]{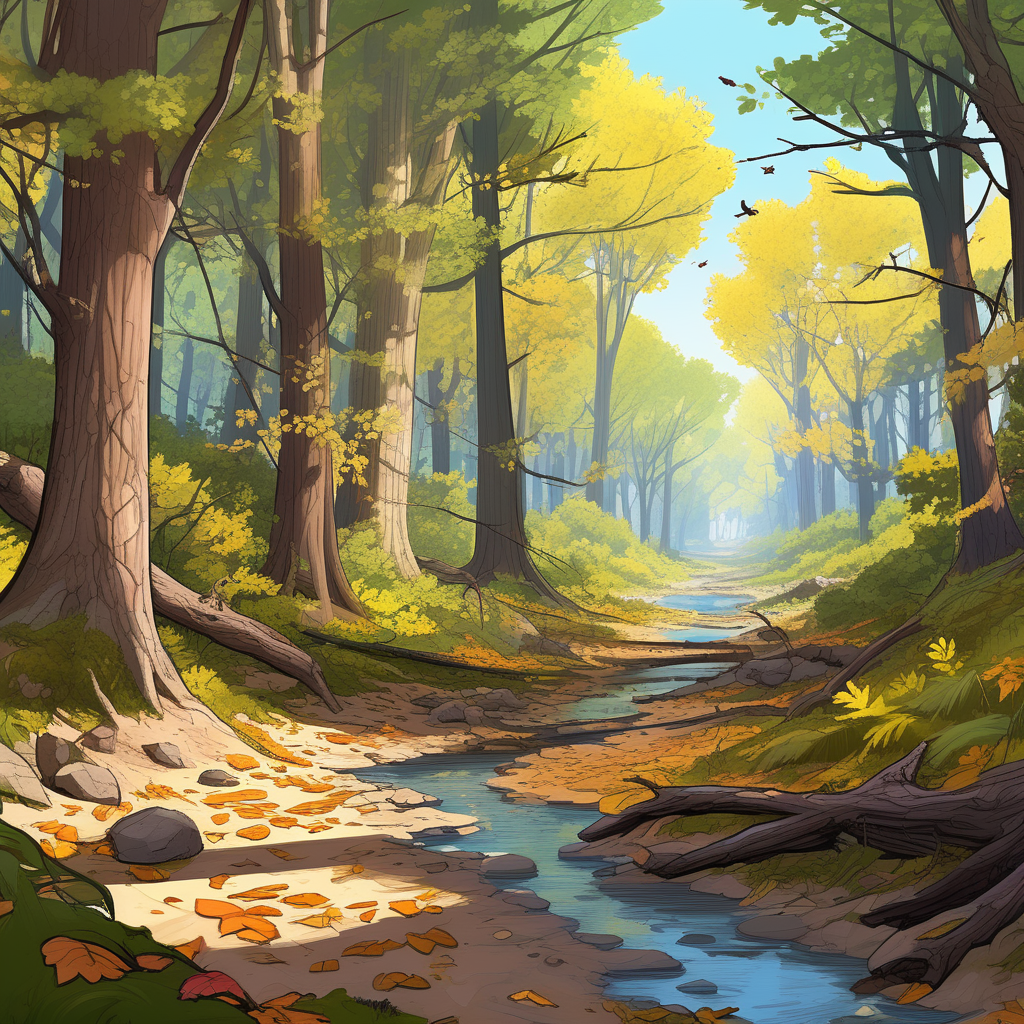} 
    \end{minipage}
    \hfill 
    \begin{minipage}[c]{0.55\linewidth}
        \tightscript{
[Visual] A dense forest with towering trees, a carpet of fallen leaves, a narrow trail, bird nests in branches, and a quiet stream.
      
[Audio] Background sounds of rustling leaves, chirping birds, and distant animal sounds create an atmosphere of being deep in the woods.}{}
    \end{minipage}
     \hfill
     
  \mbox{}
  \caption{\label{fig:consistency2} Consistency across modalities. Our framework ensure the coherence between visual and audio elements by aligning their text descriptions.}
  \vspace{-20pt}
\end{figure}

Additionally, our framework ensures consistency across modalities, as visual and audio elements together create an emotive storytelling experience. In this manuscript, the generated auditory descriptions of scenes are paired with corresponding visualizations in Fig.\ref{fig:consistency2}, illustrating how our framework maintains alignment between visual and audio styles.

\subsection{Scene interactivity}

StoryAgent can integrate detailed scene information into digital storytelling, creating scene interactivity. We focus on two common practices in digital storytelling for children, effectively engaging young audiences by emphasizing narrative elements.

\begin{figure}[htb]
  \centering
 \mbox{}
     \hfill
      \includegraphics[width=.45\linewidth]{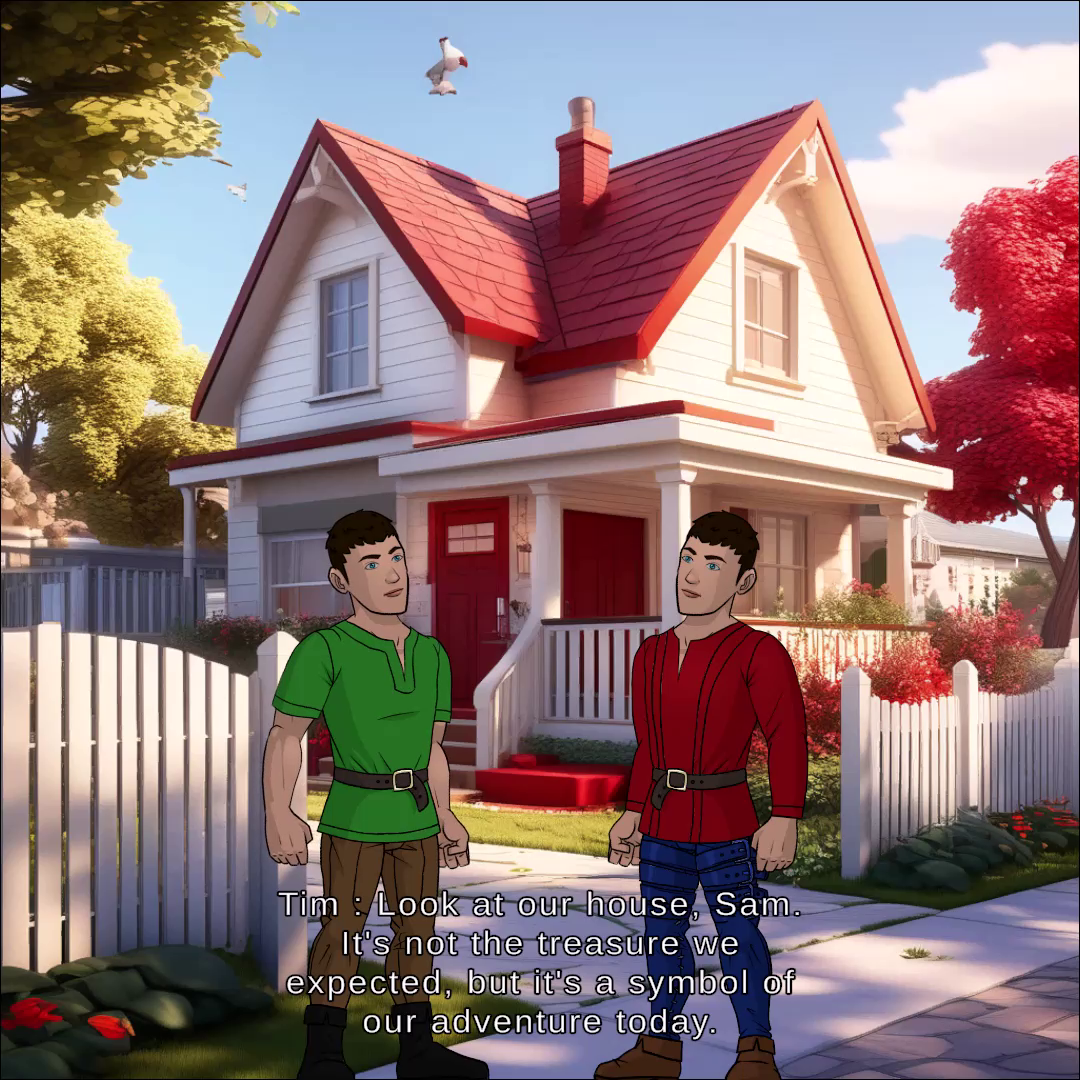}
      \hfill
      \includegraphics[width=.45\linewidth]{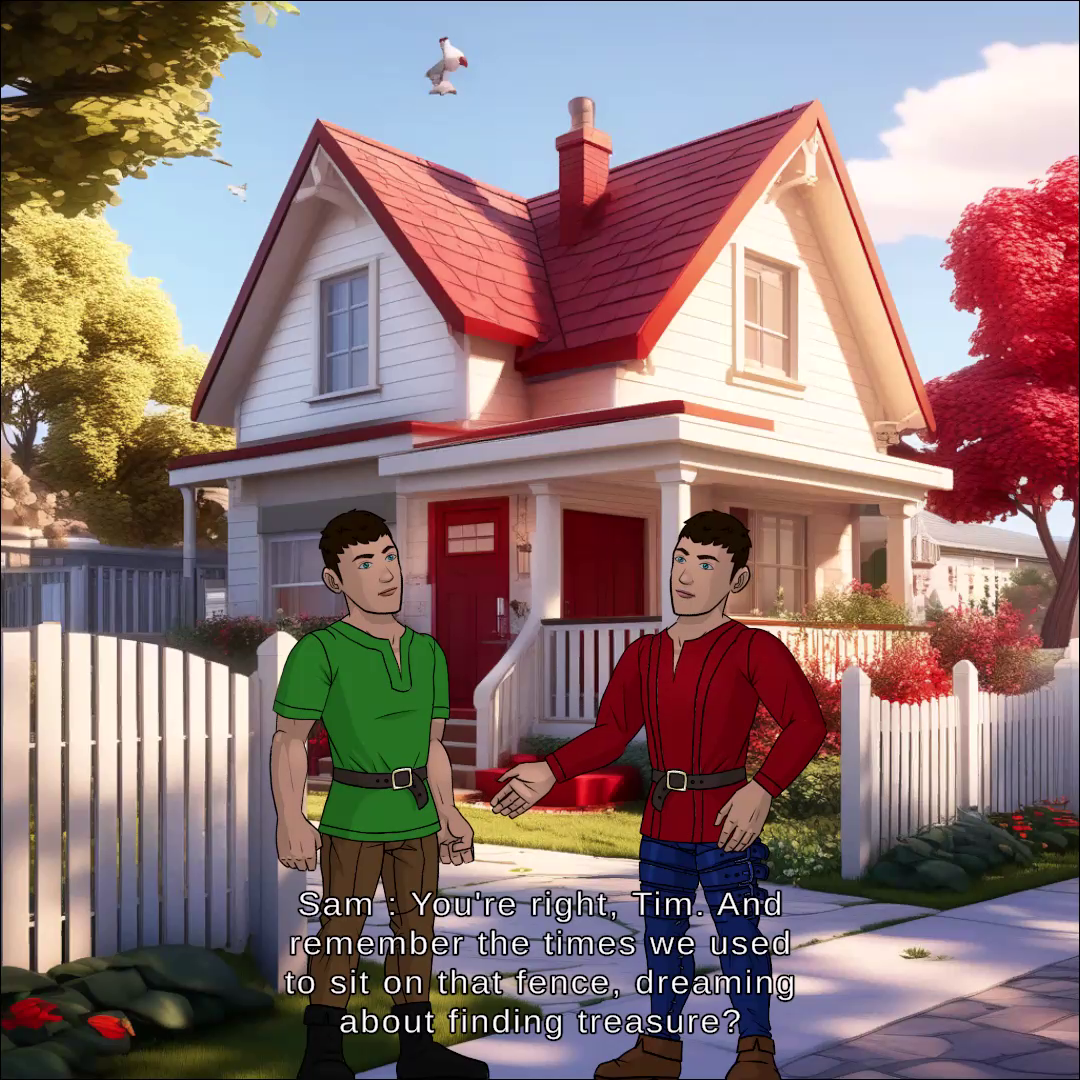}
      \hfill 
  \mbox{}
  \caption{\label{fig:scene_understand1} Character dialogue with scene understanding. (left) "Look at our house Sam. It is not the treasure we expected, but it's a symbol of our adventure today" (right) "You are right Tim. And remember the times we used to sit on that fence, dreaming about finding treasure?"}
\end{figure}

Firstly, the interactions take place at the semantic level. When the story unfolds, characters will refer to the scene objects in their dialogue, maintaining the story's coherence at the same time. In Fig.\ref{fig:scene_understand1}, the characters are talking about their childhood memories, while raising a fence in the background as a reference.

\begin{figure}
  \centering
 \mbox{}
     \hfill
      \includegraphics[width=.45\linewidth]{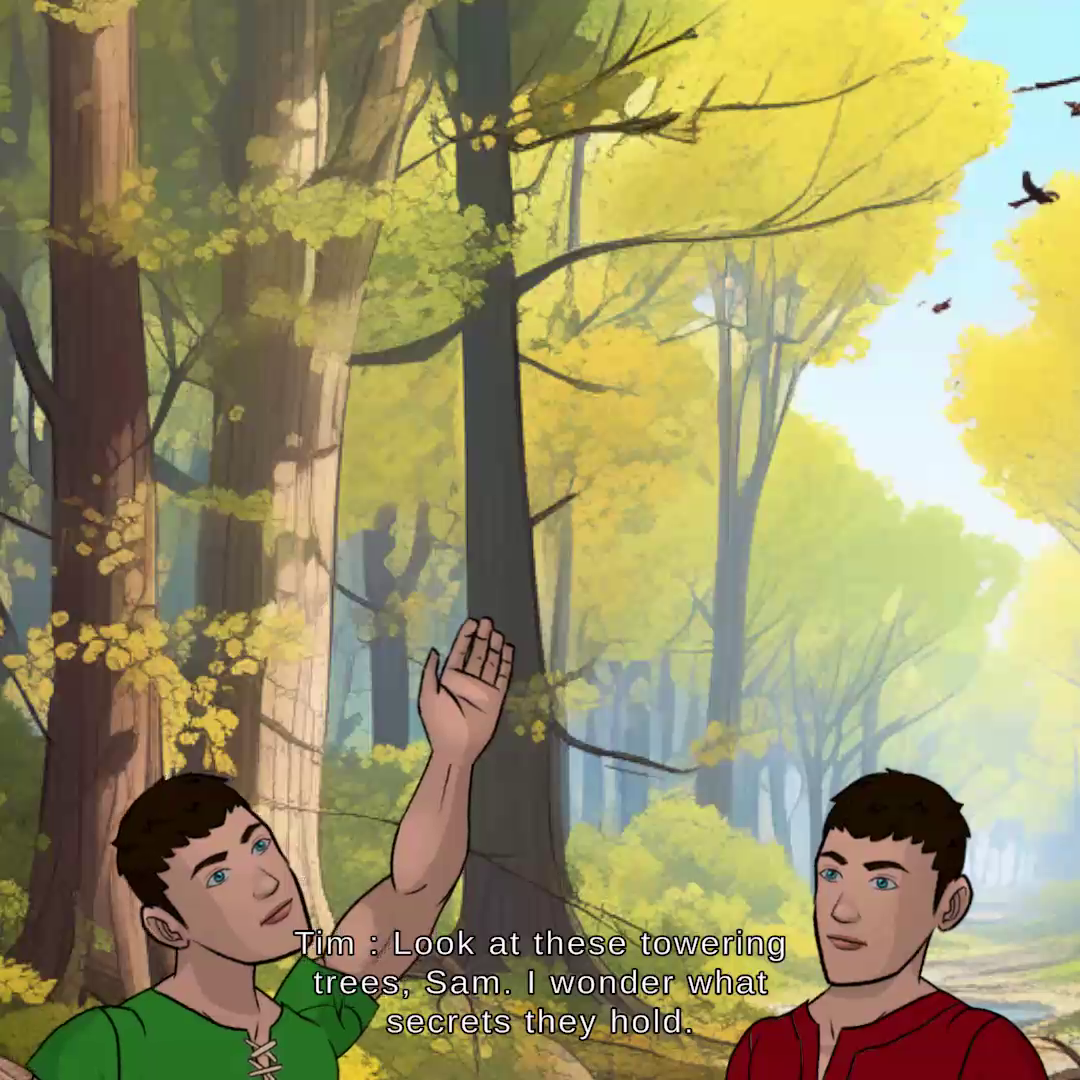}
      \hfill
      \includegraphics[width=.45\linewidth]{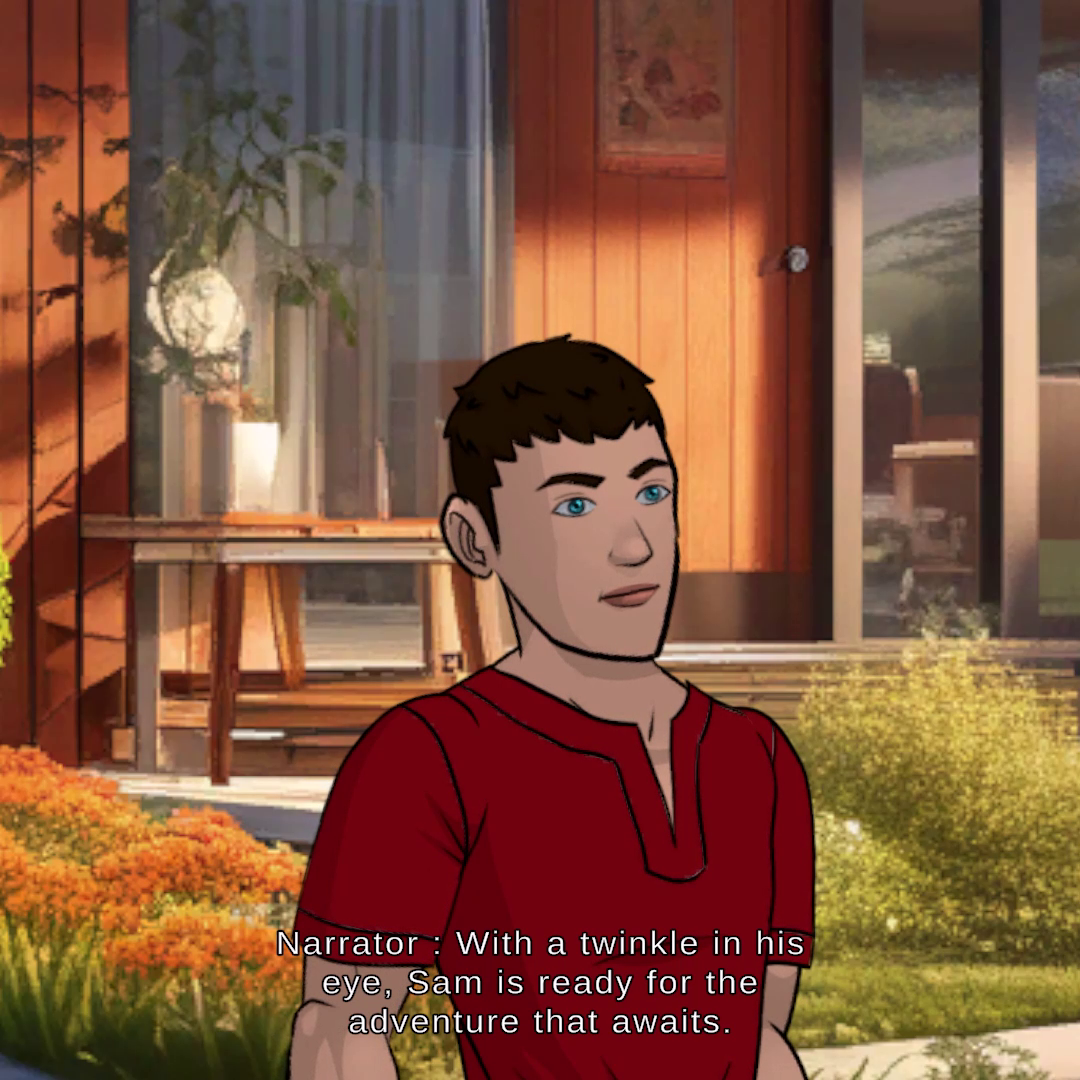}
      \hfill 
  \mbox{}
  \caption{\label{fig:scene understanding 2} Camera shoot and character gesture with scene understanding. (left) The camera shifts to focus on the trees while simultaneously, the character points at them as he discusses them. (right) The camera zooms in on the character's face while the narrator describes his eyes.}
\end{figure}

The interactivity is also enhanced by the synchronous interplay between the dialogue, character gestures, and cinematography. As characters discuss an object within the scene, not only does the camera shift focus to and zoom in on the object for detailed visualization, but the characters also direct their gazes and gestures toward it. This coordination makes digital storytelling more accessible and engaging for children, as detailed in Fig. \ref{fig:scene understanding 2}.

\begin{figure*}[htb]
  \centering
  \begin{adjustbox}{max width=\textwidth}
 \begin{tabular}{cccc}
&Base Story&Settings Intervened&Character and Settings Intervened\\
\tabtextvar{Event 1: Decipher the code}{.15}&\includegraphics[width=.25\linewidth]{figs/base-story__3_.png}&\includegraphics[width=.25\linewidth]{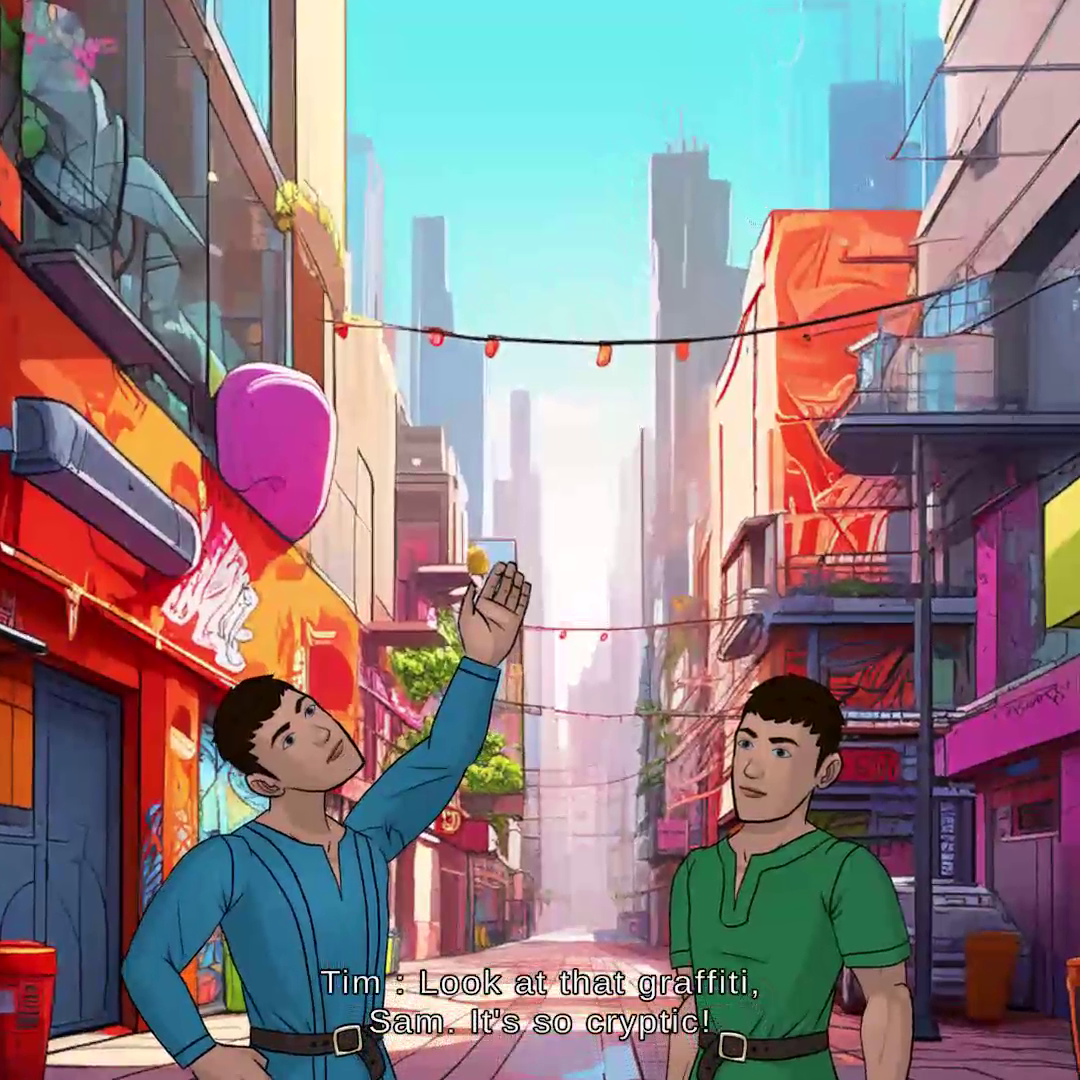}&\includegraphics[width=.25\linewidth]{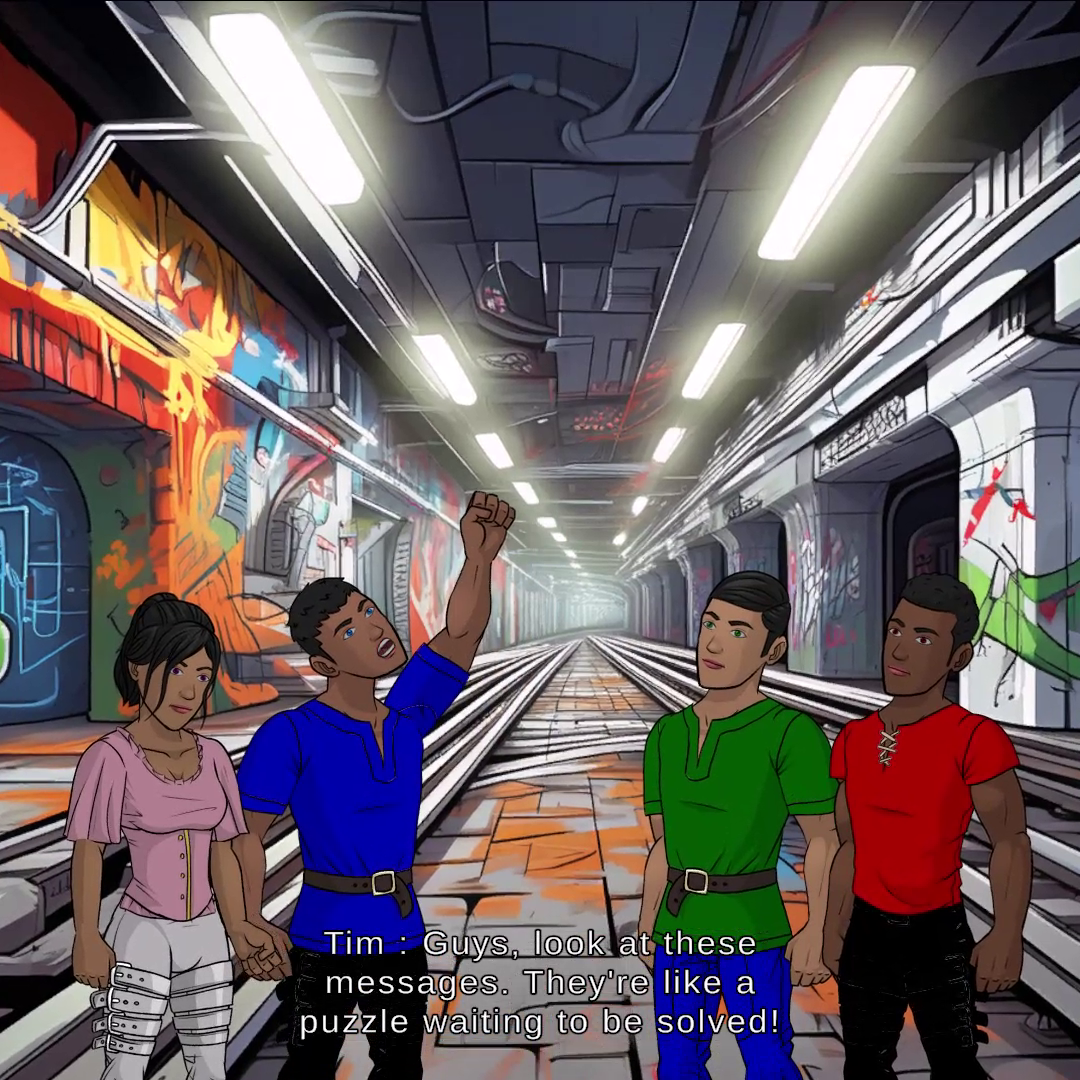}\\
&\tabtext{These carvings..they might be a code...}&\tabtext{Look at that graffiti, Sam. It's so cryptic!}&\tabtext{Guys, look at these messages. They are like a puzzle...} \\
\tabtextvar{Event 2: Navigate back to origin}{.15}&\includegraphics[width=.25\linewidth]{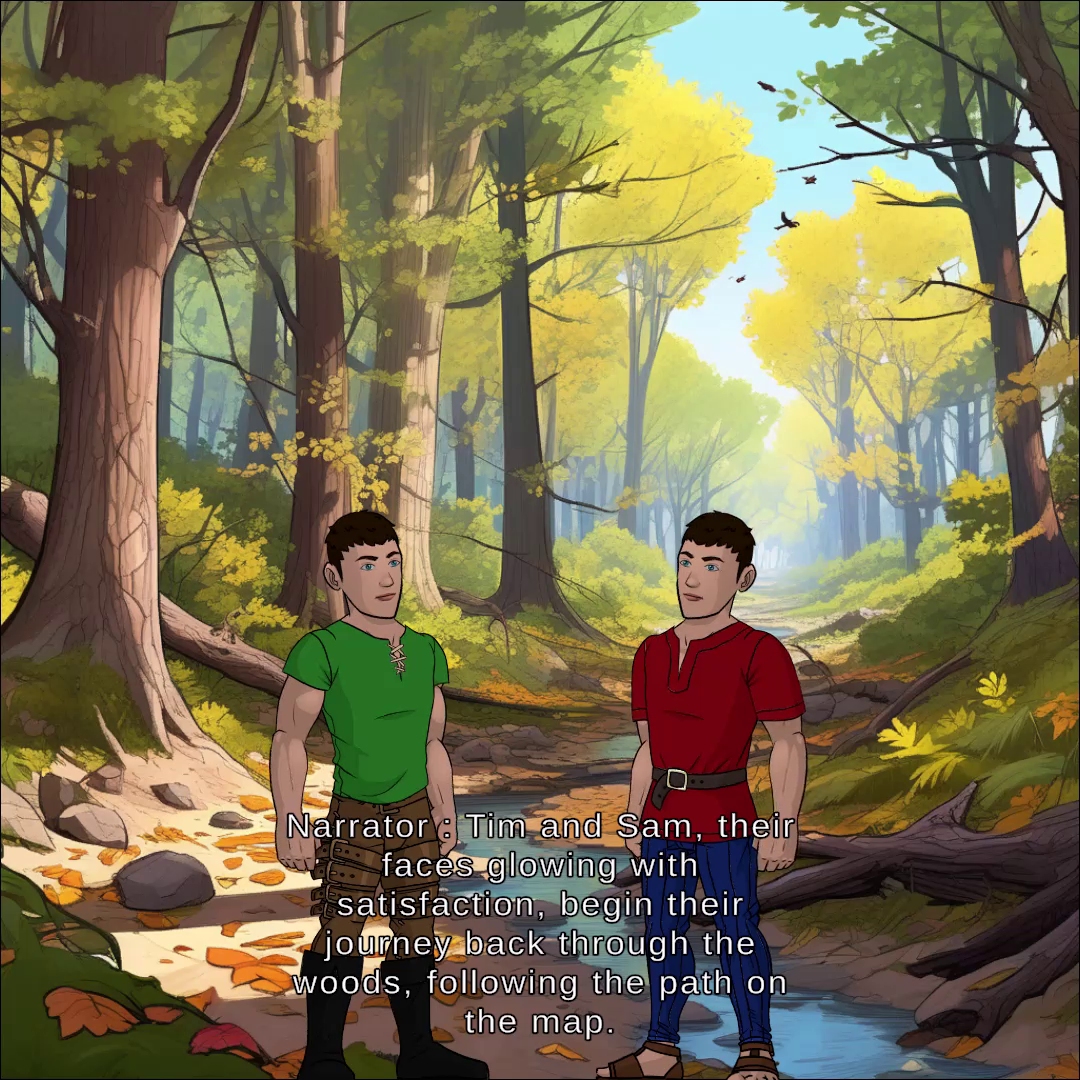}&\includegraphics[width=.25\linewidth]{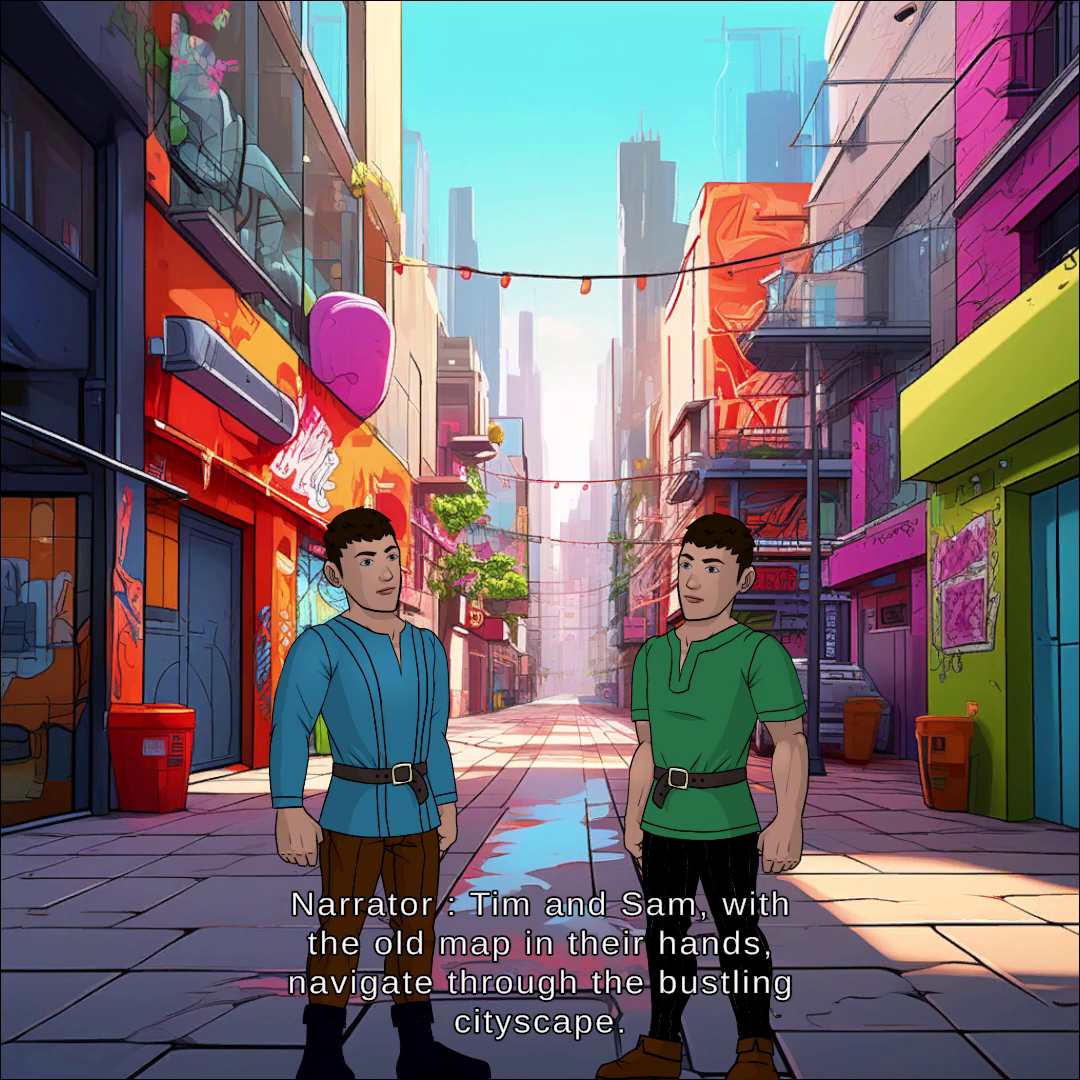}&\includegraphics[width=.25\linewidth]{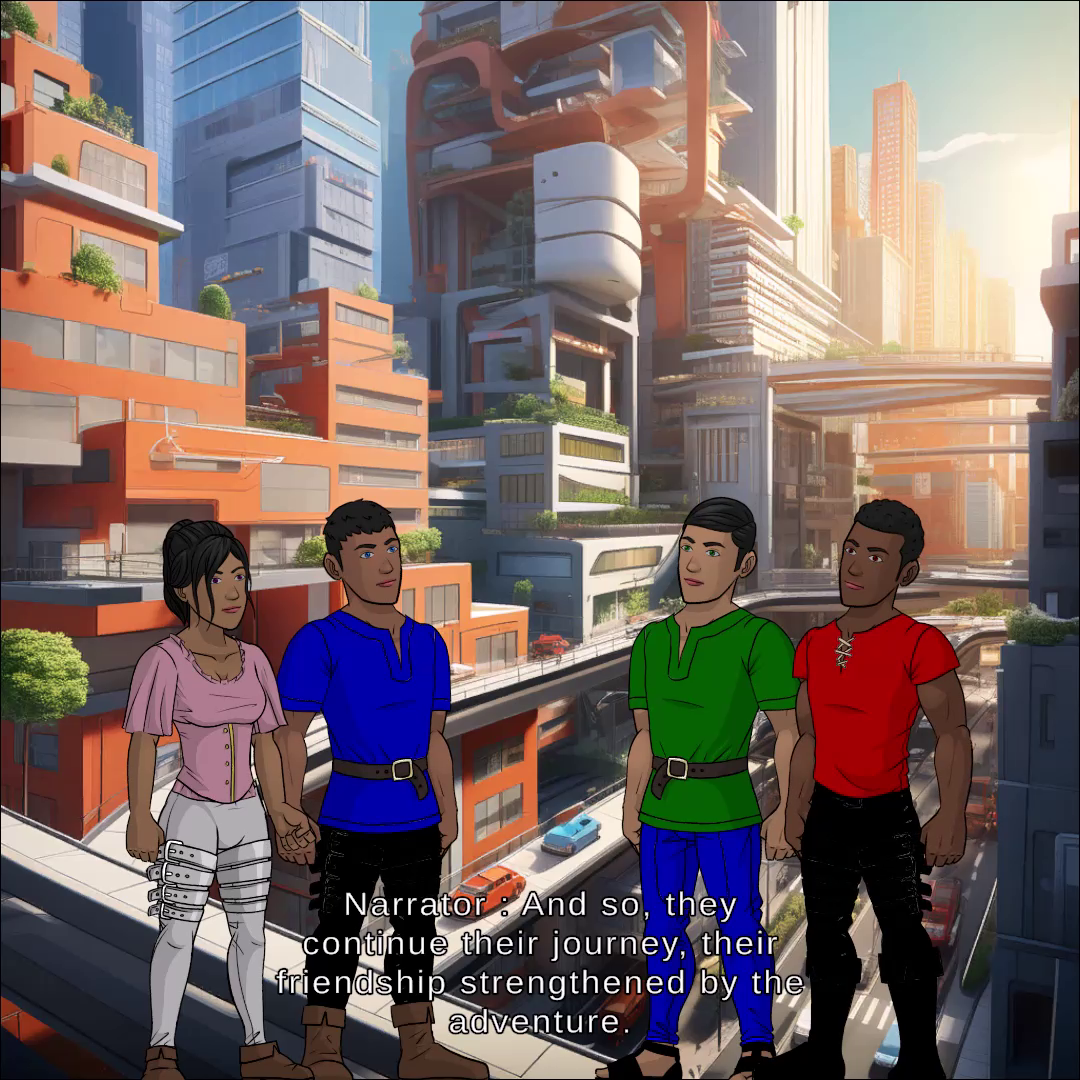}\\
&\tabtext{Tim and Sam...begin their journey back through the woods}&\tabtext{Tim and Sam...navigate through the bustling cityscape}&\tabtext{They continue their journey...}\\
\tabtextvar{Event 3: Treasure is the adventure}{.15}&\includegraphics[width=.25\linewidth]{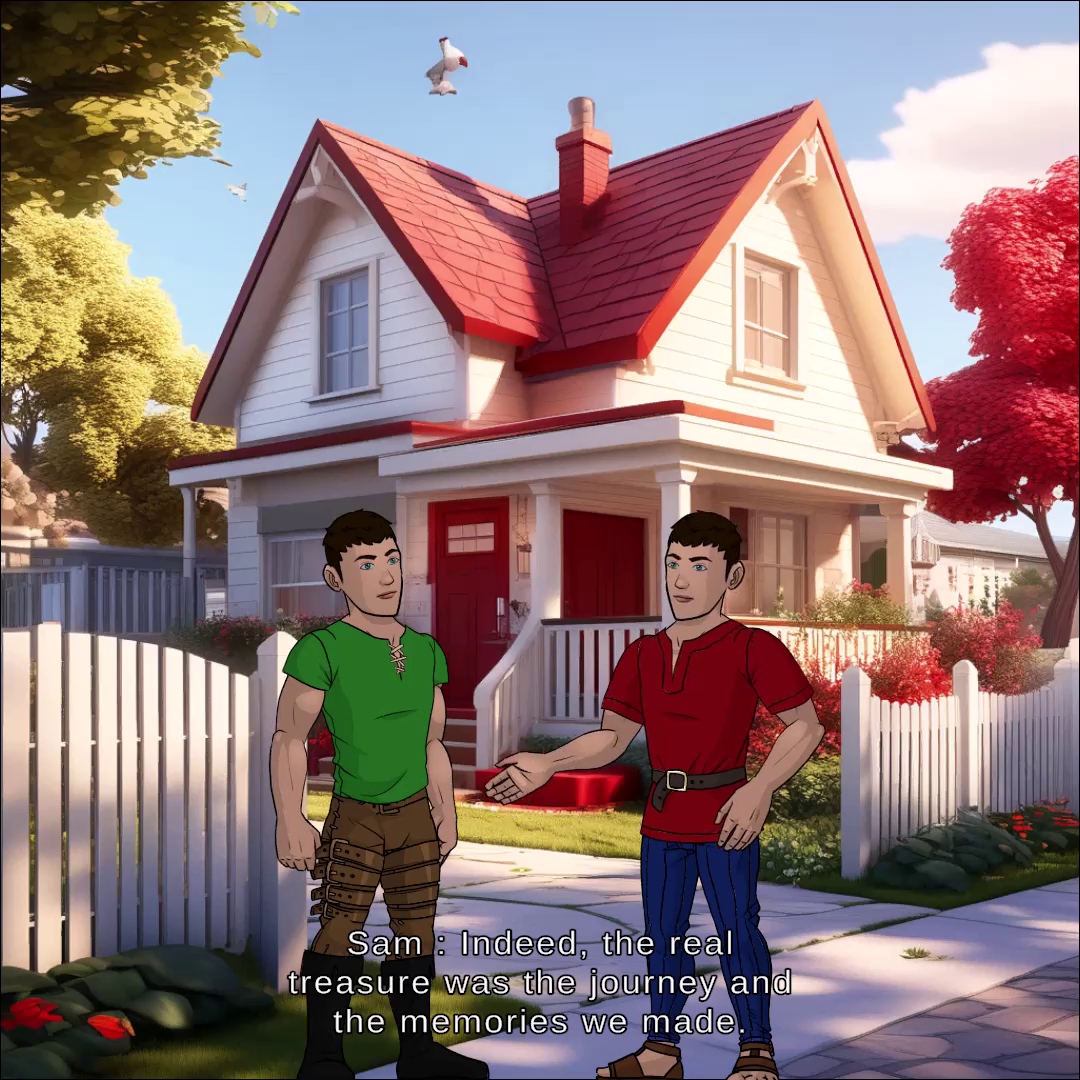}&\includegraphics[width=.25\linewidth]{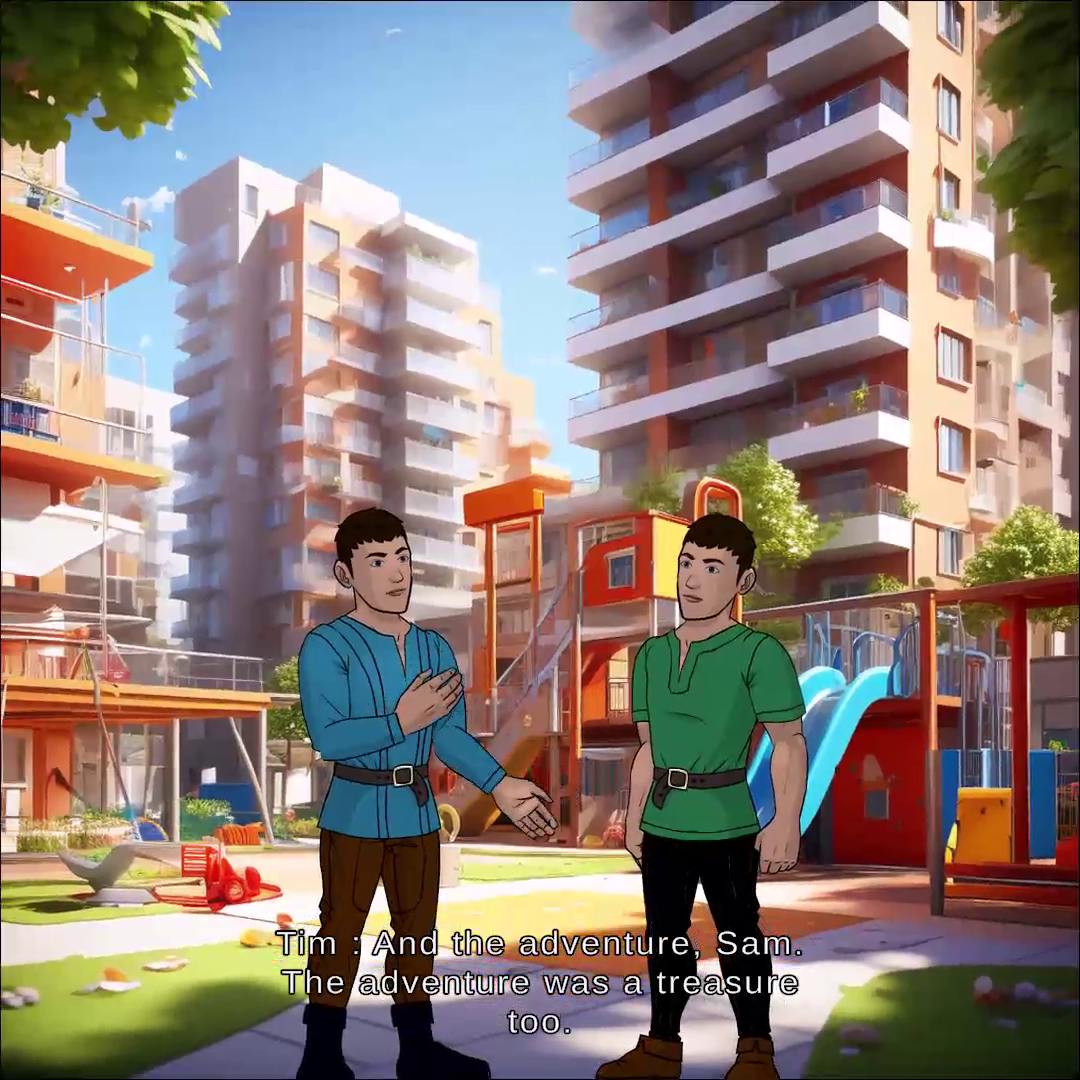}&\includegraphics[width=.25\linewidth]{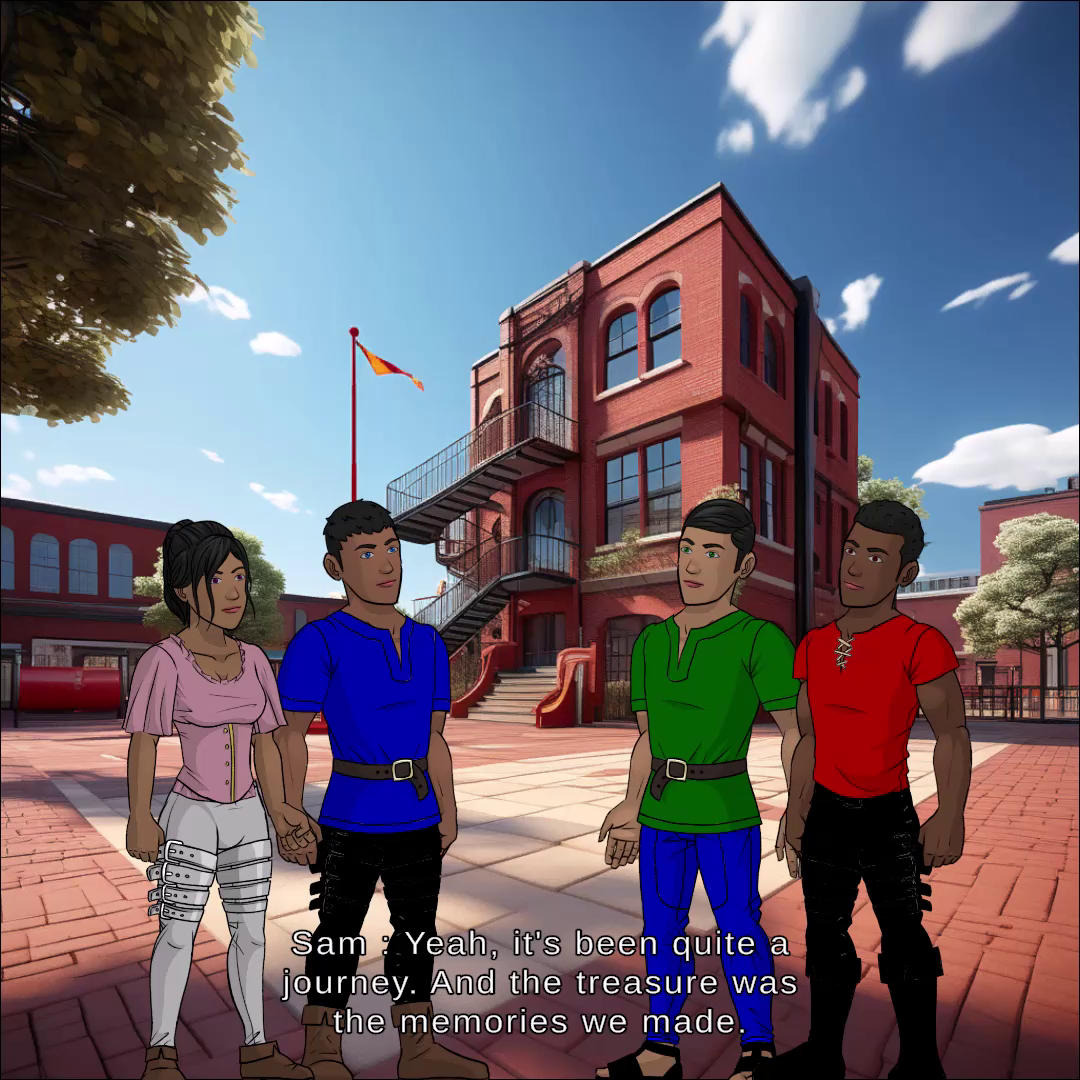}\\
&\tabtext{Indeed, the real treasure was the journey...}&\tabtext{And the adventure, Sam. The adventure is a treasure, too.} &\tabtext{...And the treasure was the memories we made.}\\
\end{tabular}
\end{adjustbox}
  \caption{\label{fig:interve} Story outcomes (screenshots and speech) with agent-wise intervention. The original base story is shown in the first column, while a variant on world settings is shown in the second. The last column is the variant where both world settings and character settings are intervened. The results show a strong complaint to the original plot, as all outcomes show the same key events (shown in rows). However, the diversity of the intervention modality is ensured at the same time.}
\end{figure*}

\subsection{Agent-wise human intervention}
\label{sec:intervention}

Our framework's coarse-to-fine approach to generating digital narratives lends itself to flexible human intervention because all agents that are affected by the intervention are strictly downstream.
Below, we demonstrate this flexibility by altering a base story generation outcome at various locations within the pipeline, each corresponding to a different modality in the storytelling process.

First, our framework demonstrates its capability to adapt writing styles while preserving the story's outline. Here we directly instruct a downstream agent team to compose the dialogue in the screenplay as rhyming couplets. This allowed us to re-use the upstream content of the base story for an easier comparison.

\noindent \texttt{Base Story Script}:

\storyscript{NARRATOR: In a quiet neighborhood, we meet Tim, a curious boy with a thirst for adventure.

TIM: I've heard the rumors about the woods, but I can't help but want to explore it.

NARRATOR: Despite the rumors, Tim's curiosity is not deterred.

TIM: No matter how far I go, I know this house will always be my safe haven.}{}

\noindent \texttt{Screenplay-Intervened Story Script}:

\storyscript{
NARRATOR: In a cozy home, lived a boy named Tim, his spirit was adventurous, his curiosity never dim.

TIM: I've heard of the woods, so haunted and deep, I must explore, while the world's asleep.

NARRATOR: With a heart full of courage, and a mind full of wonder, into the woods, Tim decided to wander.

TIM: No matter how far, into the woods I delve, this house will always be my safe haven, my safe shelf.
}{}

Next, we demonstrate more complex interventions with larger-scale impacts on the narrative generation process (i.e., downstream agents). This was achieved by using the story arc of the base story as input in the framework with an additional instruction to preserve some elements of the story while changing others. Unlike the screenplay intervention, where character appearances were directly copied, this intervention fully regenerated the characters, resulting in different-colored clothing.

Figure~\ref{fig:consistency} shows two such interventions: one where the characters and plot are fixed, but the setting is changed (\textbf{S}-intervention), and another where the plot is fixed, but the characters and setting are changed (\textbf{CS}-intervention). Between both interventions, we observe that the cryptic carvings in the forest of the base story is reflected in the cryptic graffiti in the \textbf{S}-intervention and the cryptic etchings in the \textbf{CS}-intervention (Event 1). Each image corresponding to Event 1 also shows that the story is making use of affordances to have characters look and point at the cryptic imagery. In Events 2 and 3, we see that each story's plot leads the characters on a circuitous adventure back to where they started and they share the same moral of appreciating the adventure. In the \textbf{CS}-intervention, the dialogue that was originally distributed between two characters has been distributed among four characters, meaning that the characters have not only been added but are being fully utilized. Unedited videos of the original story and the three interventions can be found in the Supplementary Materials.

\section{Discussions}

In this paper, we introduce the "StoryAgent"  framework, which utilizes text as a central medium to organize existing generative models to produce long-duration digital storytelling.
While the state-of-the-art prior works excel in the quality of short animation clips, our work is designed to address two principal hurdles in storytelling: long-duration alignment between modalities and scene interactivity.

In addition, we argue such a framework offers more flexibility in real-world digital storytelling production: First, the rapid emergence of new generative models and tools for various modalities necessitates their swift applications into a larger pipeline at minimal cost. Our model-independent framework facilitates a plug-and-play approach, allowing the latest research advancements to be easily incorporated. This ensures that the performance of our framework continuously benefits from ongoing research developments.  Second, by replacing generative assets with human-crafted ones, our framework addresses the specific limitations faced by indie developers in certain modalities. For example, graphic artists can use our pipeline to integrate music into their digital stories, whereas musicians can employ it to generate visual representations of their audio works. Finally, the text-based nature of our framework permits human artists to readily adjust intermediate products, thereby enhancing the controllability of the production process.

However, we identify several limitations in our study as follows:
Firstly, while our approach has achieved broad coverage of story components, the narrative presentation—encompassing duration, order, and style of storytelling (collectively referred to as 'discourse' in \cite{storysurvey})—is only minimally explored.
Secondly, we acknowledge that the current generation results are constrained by the performance and capabilities of state-of-the-art singular-modality models. This limitation affects the alignment between the plots and the downstream components, potentially impacting the overall coherence and immersion of the generated results.
Lastly, our experimentation with scene interactivity is limited to a small number of scenarios. Future research should explore more complex interaction modes, such as enabling characters to pick up items or alter the status of objects within the environment, to enhance the interactive experience in 2D art styles.

\bibliography{main}

\clearpage
\newpage
\section{Appendix}

\subsection{Prompts in story cluster}
\label{sec:prompt_story_cluster}

For their system prompts, all LLM agents within the story cluster utilize one of two templates provided below. Each template is tailored to a specific agent by filling in details for the agent role, constraint, and output format. Initially, we present the templates in full and subsequently, we focus exclusively on the aforementioned three components, which are illustrated as placeholders within each template.

\noindent Expert Agent

\begin{lstlisting}[language=text,numbers=none,basicstyle=\ttfamily,]
{upstream_contents}

Expert {agent_name}.
Role:
{agent_role}

Constraints:
{agent_constraints}

ONLY OUTPUT A JSON FILE.
The expert must include an explanation of how each constraint is followed.

Format:
{agent_output}
\end{lstlisting}

\noindent Critic Agent

\begin{lstlisting}[language=text,numbers=none,basicstyle=\ttfamily,]
{upstream_contents}

Critic {agent_name}. 
Check that the expert is performing its role and adhering to the following constraints.

Role:
{agent_role}

Constraints:
{agent_constraints}

For every single constraint, output a constraint score between 0.0 and 1.0 (inclusive) that judges the quality of Expert's output and explain your reasoning. Do not omit any constraints. Be very strict.

Also ensure that the expert's output adheres perfectly to the following format and adds nothing else:

{agent_output}

Output a format score between 0.0 and 1.0 (inclusive) for whether the output is strictly in JSON format. 1.0 means that the format is perfect. Be very strict.
DO NOT FORGET the leading and trailing "```json" and "```".
DO NOT CREATE unterminated strings.

If the expert has not completed all of its turns, its turn constraint score must be 0.0.

If and only if the constraint and format scores are all equal to 1.0 (meaning there are no suggestions), you must print TERMINATE.
Otherwise, explain to the Expert exactly why the score is less than 1.0, make suggestions on how they can improve their score and DO NOT PRINT TERMINATE IN ANY WAY.
DO NOT SAY "we cannot print TERMINATE".

DO NOT ASK QUESTIONS.
\end{lstlisting}

\subsubsection{Story arc team}
Agent role
\begin{lstlisting}[language=text,numbers=none,basicstyle=\ttfamily,]
In your first turn, use the story input to output a general story arc that describes the exposition, rising action, climax, falling action, and resolution.
In your second turn, expand upon anything vague. E.g., if challenges/puzzles/mysteries are mentioned, explain what they are specifically.
\end{lstlisting}

\noindent Agent constraint
\begin{lstlisting}[language=text,numbers=none,basicstyle=\ttfamily,]
1. Each character must be a singular human.
2. There must be no non-human characters.
3. Characters cannot interact with non-human characters in the story.
4. There must be multiple main characters that talk to each other throughout the story.
5. Keep the story simple.
6. Do not be vague (e.g., "they solve a puzzle", "they encounter a mystery"), be specific!
7. It must be possible to animate the story with characters only speaking and walking. They cannot climb, jump, or hold anything.
8. Make sure that with each change, the motifs in the story stay consistent.
\end{lstlisting}

\noindent Agent output
\begin{lstlisting}[language=text,numbers=none,basicstyle=\ttfamily,]
The output should be a markdown code snippet formatted in the following schema, including the leading and trailing "```json" and "```":

```json
{
"story_arc": dict  // Story arc
{
    "exposition": str
    "rising_action": str
    "climax": str
    "falling_action": str
    "resolution": str
}
"constraints": list[str]  // Explanation for each constraint
}
```
\end{lstlisting}

\subsubsection{Characters team}
Agent role
\begin{lstlisting}[language=text,numbers=none,basicstyle=\ttfamily,]
For each character, describe their name, gender, age, personality, beliefs, motivations, development, and physical description.
The physical description must describe the fabric, color, texture, and accessories of the tunic, pants, and boots that the character is wearing.

In your first turn, for each part of the story arc (exposition, rising action, climax, falling action, and resolution) create a list of every single human mentioned directly or indirectly with empty attributes.
In your second turn, collate these lists and populate every character's attributes with broad details that are consistent with the story arc.
In your third turn, make attributes much more detailed.

\end{lstlisting}

\noindent Agent constraint
\begin{lstlisting}[language=text,numbers=none,basicstyle=\ttfamily,]
1. Every human mentioned in the story arc must be in the character list.
2. There must be more than 1 character listed.

3. Every character must be human.
4. Every character name must be singular, e.g., an entry cannot be named "villagers" or "siblings" because these are plural.

5. There must be a Narrator with empty attributes in its physical description.

6. Each character's attributes must be non-empty, unambiguous, and detailed.

\end{lstlisting}

\noindent Agent output
\begin{lstlisting}[language=text,numbers=none,basicstyle=\ttfamily,]
The output should be a markdown code snippet formatted in the following schema, including the leading and trailing "```json" and "```":

```json
{
"story_characters": list[dict]  // List of story characters
{
"name": str  // Character name
"gender": str  // Character gender: "Male" or "Female"
"age": int  // Character age
"personality": str  // Character's personality
"beliefs": str  // Character's beliefs
"motivations": str  // Character's motivations
"development": str  // How the character develops throughout the story
"physical_description": dict[str]  // Character's physical description
    {
        "tunic": str  // Description of tunic
        "pants": str  // Description of pants
        "boots": str  // Description of boots
    }
}
"constraints": list[str]  // Explanation for each constraint
}
```
\end{lstlisting}

\subsubsection{Settings team}
\paragraph{Stage 1}

Agent role
\begin{lstlisting}[language=text,numbers=none,basicstyle=\ttfamily,]
Create a hierarchical graph of settings directly or indirectly mentioned in the story arc, where each node is a setting consists of its name, its parent setting, a list of its children settings, and story-related information.
The parent of the current setting must be a larger area that the current setting must be fully contained within.
A child of the current setting must be a smaller area that is fully contained within the current setting.
All children within a setting should be navigable between each other without leaving the setting.

The first setting should be named World and all other settings are contained within this setting.
The World setting's parent must be an empty string.

In your first turn, create the setting hierarchy and omit "is_outside" attributes.
In your second turn, for each setting that has an interior enter (e.g., a house), create appropriate children settings for the inside that are relevant to the story arc.
In your third turn, fill in the "is_outside" attribute for each setting.

\end{lstlisting}

\noindent Agent constraint
\begin{lstlisting}[language=text,numbers=none,basicstyle=\ttfamily,]
1. Every single setting mentioned directly or indirectly in the story arc (both large and small) must be included.
2. For every setting that has an interior (e.g., a house), there must be at least one child setting so that characters can enter.
3. The difference in scale between a parent setting and child setting must be gradual. Create intermediate settings is this is not the case.
4. If a setting is a building (e.g., a house, mansion, shed, etc.), its is_outside attribute must be true, but its children settings (rooms) must have is_outside = false.
5. All setting names must be unique.
6. Make sure the expert completes 3 turns.
\end{lstlisting}

\noindent Agent output
\begin{lstlisting}[language=text,numbers=none,basicstyle=\ttfamily,]
The output should be a markdown code snippet formatted in the following schema, including the leading and trailing "```json" and "```":

```json
{
"story_settings": list[dict]  // List of settings
{
    "name": str  // Name of the setting
    "is_outside": bool  // Indicates whether the setting is inside or outside.
    "parent": str  // Name of parent setting or empty string if none
    "children": list[str]  // List of children settings
}
"constraints": list[str]  // Explanation for each constraint
}
```
\end{lstlisting}

\paragraph{Stage 2}

Agent role
\begin{lstlisting}[language=text,numbers=none,basicstyle=\ttfamily,]
For each setting, add a visual prompt that summarizes its physical state. Include the visible square footage of walkable area and the visual style that all settings should have (e.g., time period).
\end{lstlisting}

\noindent Agent constraint
\begin{lstlisting}[language=text,numbers=none,basicstyle=\ttfamily,]
1. The visual prompt should not include any information about the characters.
2. The visual prompt must describe at least 5 interesting objects (big or small) in the setting.
3. The visual prompt should be written in prose and be concise and literal and not figurative.
4. The visual prompt must be UNDER 60 words long. Do not waste any words!
5. If a setting has is_outdoors = true, the square footage should also consider the surrounding area. If the setting has is_outdoors = false, the square footage should be limited to the room.
6. If a setting has is_outdoors = true, ONLY DESCRIBE THE OUTSIDE of the setting. If the setting has is_outdoors = false, ONLY DESCRIBE THE INSIDE.
7. The square_footage of each setting must be at least 100. Otherwise, add details about the parent setting into the visual prompt and update square_footage accordingly.
8. DO NOT OMIT ANY SETTINGS.

\end{lstlisting}

\noindent Agent output
\begin{lstlisting}[language=text,numbers=none,basicstyle=\ttfamily,]
The output should be a markdown code snippet formatted in the following schema, including the leading and trailing "```json" and "```":

```json
{
"story_settings": list[dict]  // List of settings
{
    "name": str  // Name of the setting
    "is_outside": bool  // Indicates whether the setting is inside or outside.
    "parent": str  // Name of parent setting or empty string if none
    "children": list[str]  // List of children settings
    "visual_prompt": str  // Description of physical state for stable diffusion prompt
    "square_footage": int  // Visible square footage of setting
}
"visual_style": str  // Visual style of all settings
"constraints": list[str]  // Explanation for each constraint
}
```
\end{lstlisting}

\subsubsection{Storybeat team}
Agent role
\begin{lstlisting}[language=text,numbers=none,basicstyle=\ttfamily,]
Each story beat has a description, a list of characters involved, a list of settings involved, time of day, and how characters and the audience should feel.

In your first turn, output a detailed list of story beats.
In your second turn, split story beats that have multiple settings into multiple story beats with single settings.

\end{lstlisting}

\noindent Agent constraint
\begin{lstlisting}[language=text,numbers=none,basicstyle=\ttfamily,]
1. There must be at least 1 story beat for each part of the story arc (exposition, rising action, climax, falling action, resolution).
2. Do not introduce new settings or characters that are not in story_characters and story_settings.
3. Each story beat's characters list must include Narrator.
4. The passage of time during a story beat should be reasonably small to minimize the amount of jumps in time.
\end{lstlisting}

\noindent Agent output
\begin{lstlisting}[language=text,numbers=none,basicstyle=\ttfamily,]
The output should be a markdown code snippet formatted in the following schema, including the leading and trailing "```json" and "```":

```json
{
"story_beats": list[dict]  // Story beats
{
    "description": str  // Description of what happens in the story beat
    "settings": list[str]  // List of setting used in the story beat
    "characters": list[str]  // List of characters present in the story beat
    "time_passage": str  // How much time passes during the story beat
    "character_feelings": str  // Describe how characters feel during the story beat
    "audience_feelings": str  // Describe how the audience should feel during the story beat
}
"constraints": list[str]  // Explanation for each constraint
}
```
\end{lstlisting}

\subsubsection{Setting affordance team}
\label{app:affordance_agent}
\paragraph{Stage 1}

Agent role
\begin{lstlisting}[language=text,numbers=none,basicstyle=\ttfamily,]
setting_objects contains a list of objects detected in that image, their bounding boxes, and their depths. KEEP IN MIND that bounding boxes can be mislabeled or false positives and 1 object can be detected as multiple bounding boxes.

In your first turn, add each entry in setting_objects to object_list and use the bboxes (detected bounding boxes) and bbox_depths (average depths of the bounding boxes) to describe the configuration of the objects in high detail. Output empty strings for spatial_relations and evidence.
In your second turn, add the spatial_relations describing how the configurations are related BETWEEN different types of objects in high detail. Output empty strings for evidence.

\end{lstlisting}

\noindent Agent constraint
\begin{lstlisting}[language=text,numbers=none,basicstyle=\ttfamily,]
1. The spatial_relations for each object should be as specific as possible!
2. Make sure the expert completes 2 turns.
\end{lstlisting}

\noindent Agent output
\begin{lstlisting}[language=text,numbers=none,basicstyle=\ttfamily,]
The output should be a markdown code snippet formatted in the following schema, including the leading and trailing "```json" and "```":

```json
{
"setting_affordances": list[dict]  // List of affordances per setting
{
    "name": str  // Name of the setting
    "object_list": list[dict]  // List of affordances
        {
            "object_name": str  // Object name
            "configuration": str  // Description of how the bounding boxes of the object are arranged in space
            "spatial_relations": str  // Description of how the object is configured with respect to other objects.
        }
}
"constraints": list[str]  // Explanation for each constraint
}
```
\end{lstlisting}

\paragraph{Stage 2}

Agent role
\begin{lstlisting}[language=text,numbers=none,basicstyle=\ttfamily,]
setting_prompt is the prompt used to generate an image.
setting_objects contains a list of objects detected in that image, their bounding boxes, and their depths. KEEP IN MIND that some bounding boxes are false positives and 1 object can be detected as multiple bounding boxes.
setting_caption describes what is visible in the image.
setting_beats lists what characters will do in the setting.

In your first turn, add the evidence that the objects appear in the image using the setting_prompt and the setting_caption and rate that evidence from 0 to 1. Leave the affordances list completely empty.
	Even if an object is not explicitly mentioned in the prompt or caption, if it is highly related to the context, its evidence should be high.
	An evidence rating below 0.5 means that the object was falsely detected.
	Keep in mind that the object detection system can misinterpret the image, the captioning system can misinterpret the image, and the image generation can misinterpret the prompt.
	You can also use the bounding boxes as evidence, e.g., if compared to the bounding box of another object, the current object is too small/large (depth must also be accounted for).
In your second turn, use the story beats to populate the affordances with things that characters can say that are relevant to the beat.
	The dialogue MUST NOT say that an object looks a certain way.
	Score the rating of the affordance (from 0 to 1) based on how much it aligns with character_feelings and audience_feelings.
	0 means that the dialogue is not critical to the story.
	1 means that the dialogue is critical to the story.
\end{lstlisting}

\noindent Agent constraint
\begin{lstlisting}[language=text,numbers=none,basicstyle=\ttfamily,]
1. Dialogue in the affordances must not describe the appearance of the objects.
2. The affordances list must not be empty for any object.
3. Do not omit any objects.
4. Do not output configuration or spatial_relations.
5. The affordances list must be as long as the setting_beats list.
6. Make sure the expert completes 2 turns.
\end{lstlisting}

\noindent Agent output
\label{app:affordance_output}
\begin{lstlisting}[language=text,numbers=none,basicstyle=\ttfamily,]
The output should be a markdown code snippet formatted in the following schema, including the leading and trailing "```json" and "```":

```json
{
"setting_affordances": dict  // List of affordances per setting
{
"name": str  // Name of the setting
"object_list": list[dict]  // List of affordances
{
    "object_name": str  // Object name
    "configuration": str  // Description of how the bounding boxes of the object are arranged in space
    "spatial_relations": str  // Description of how the object is configured with respect to other objects.
    "evidence": str  // Evidence supporting the object's presence in the image
    "evidence_rating": float  // Value between 0 and 1 judging the amount of evidence
    "affordances": dict  // List of things that characters can say about the object in each relevant story beat.
    {
        "dialogue": str  // Character dialogue
        "reason": str  // Reasoning based on story beat
        "rating": float  // Value between 0 and 1 judging importance to story beat
    }
}
}
"constraints": list[str]  // Explanation for each constraint
}
```
\end{lstlisting}

\subsubsection{Story scene team}
\label{app:story_scene_agent}
Agent role
\begin{lstlisting}[language=text,numbers=none,basicstyle=\ttfamily,]
A scene is a set of instructions for a 2D animator.
The characters parameter is a list of all characters that are present during the scene, meaning that they have dialogue or are visible.
The setting parameter is depicted as a static background image that characters walk in front of.
	Characters are only allowed to talk, so they cannot perform any other actions.
The objects parameter lists objects that are part of the image that characters can speak about.
	If there are no affordances, the object list MUST BE EMPTY.
	All objects must be from the setting_affordances.
The dialogue parameter describes what all characters talk about in high detail. It should not include direct quotes.
The sound_effects and music_effects parameters must be consistent with the character_feelings, audience_feelings, and setting.

Your role is to decide precisely how the story beat should be visually and auditorily portrayed through a series of scenes.

\end{lstlisting}

\noindent Agent constraint
\begin{lstlisting}[language=text,numbers=none,basicstyle=\ttfamily,]
1. Do not makes scenes that are redundant with the previous or next story beats.
2. Since characters can only talk, if the story beat requires a character to do anything else, you must convey this by having the narrator dictate it.
3. Every scene must have a sound and music effects.
4. If a scene has dialogue from characters or the Narrator, the dialogue description must describe what they say in full detail and nothing else.
5. The scene cannot include details that cannot be animated by characters. Characters cannot be said to make certain facial expressions.
6. Do not create more scenes than necessary.
7. There must be no new characters that are not in the character list.
8. The setting of the scene must be the lowest suitable child setting. E.g., if two people are talking in the house, the setting should not be house, but a child setting of the house, because house is the outside. If the people are talking directly outside the house, then the setting should be house.
9. Each character must have dialogue.
10. The dialogue must include at least 1 object in the setting and must blend it seamlessly into the conversation.
11. If there are no affordances, the objects list must be empty.
12. All objects must be from the setting_affordances
\end{lstlisting}

\noindent Agent output
\begin{lstlisting}[language=text,numbers=none,basicstyle=\ttfamily,]
The output should be a markdown code snippet formatted in the following schema, including the leading and trailing "```json" and "```":

```json
{
"scene_count_explanation": str  // Explanation of the number of scenes
"story_scenes": list[dict]  // Story scenes
{
    "characters": list[str]  // List of present characters
    "setting": str  // Setting name
    "objects": list[str]  // List of objects that characters interact with
    "dialogue": str  // Description of character dialogue
    "sound_effects": str  // A description of background sound effects
    "music_effects": str  // A description of background music
}
"constraints": list[str]  // Explanation for each constraint
}
```
\end{lstlisting}

\subsubsection{Screenplay team}
Agent role
\begin{lstlisting}[language=text,numbers=none,basicstyle=\ttfamily,]
For the given story scene, output a sequence of actions similar to a screenplay.

At the start of a scene, all involved characters enter at the start from the left and exit at the end to the right.
Each action can either be CAMERA, SPEAK.

The CAMERA action is used to control the camera. It must include the following parameters: target, size.
	The target determines where the camera focuses: either "setting" or the name of a character.
	The size determines how zoomed in the camera should be: "close", "medium", "wide".
	If the target is "setting", the size must be "wide".

The SPEAK action should be used for all dialogue and must include the following parameters: speaker, line, listener, emotion, adverb, object.
	The speaker must be in the list of characters (including the Narrator).
	The line is the spoken dialogue.
	The listener must another character that the speaker is directly talking to or an empty string if the speaker is monologuing.
	The emotion must be either Happy, Sad, Angry, Afraid, Disgusted, Surprised, or Neutral.
	The adverb must describe how the line is spoken by the character.
	The object is empty by default unless the character is speaking about an object in setting_affordances.

Create at least 5 SPEAK actions. DO NOT BE REDUNDANT. Characters should not repeat each other.
A character's dialogue should reflect their traits and character_feelings. E.g., a child should not speak like an adult.

\end{lstlisting}

\noindent Agent constraint
\begin{lstlisting}[language=text,numbers=none,basicstyle=\ttfamily,]
1. No actions outside of CAMERA or SPEAK can be used.
2. Make sure that characters are consistent with the story beat's character_feelings.
3. Make sure that the audience would feel audience_feelings by the end of the action list.
4. You must incorporate at least one object from setting_affordances if it is not empty.
5. There must be at least 5 SPEAK actions.
\end{lstlisting}

\noindent Agent output
\begin{lstlisting}[language=text,numbers=none,basicstyle=\ttfamily,]
The output should be a markdown code snippet formatted in the following schema, including the leading and trailing "```json" and "```":

```json
{
"action_list": list[dict]  // Story beats
{
    "action_type": str  // Either CAMERA, SPEAK
    "params": dict  // The parameters of the action_type
}
}
```
\end{lstlisting}

\subsection{Prompts in asset generation cluster}

\subsubsection{Image}

Character parameterizer: agent role
\begin{lstlisting}[language=text,numbers=none,basicstyle=\ttfamily,]
Using the CC2D Guide below, output a dictionary for the body, skin details, values for each body slider, hair, facial hair, eyebrows, 
and a list of RGB values for each part in the Colors list based on the character's description.

All slider values are between 0 and 1 and default to 0.5 for a normal-sized adult.
If a slider has 1 value, it should still be returned as part of a list.
Lower values produce smaller body parts.

Provide an explanation for the choices.

Colors must only have RGB list of floats (between 0.0 and 1.0) as values and strings as keys. DO NOT ADD ANY NEW KEYS.
BLACK MUST NOT BE PITCH BLACK (0.0, 0.0, 0.0).

{
    "CC2D Guide": {
        "Body": {
            "Base 00 Female": "base female character",
            "Base 00 Male": "base male character"
        },
        "BodySliders": {
            "Body": "2 values for width/length of torso",
            "Ear": "1 value for ear size",
            "Foot": "1 value for foot size",
            "Hand": "1 value for hand size",
            "Head": "1 value for head size, but not ear size",
            "LowerArm": "2 values for width/length of lower arm",
            "LowerLeg": "2 values for width/length of lower leg",
            "Neck": "2 values for width/length of neck",
            "UpperArm": "2 values for width/length of upper arm",
            "UpperLeg": "2 values for width/length of upper leg"
        },
        "Colors": [
            "BodySkin",
            "SkinDetails",
            "Hair",
            "FacialHair",
            "Eyebrow",
            "Eyes",
            "Mouth",
            "Armor",
            "Pants",
            "Boots"
        ],
        "Eyebrow": {
            "00": "default",
            "01": "thinner",
            "02": "more arched",
            "03": "more straight",
            "04": "thicker and furrowed",
            "05": "bushy",
            "09": "eyebrow slit",
            "NULL": "none"
        },
        "FacialHair": {
            "09": "bushy sideburns",
            "10": "mutton chops",
            "11": "walrus mustache",
            "12": "large beard with walrus mustache",
            "NULL": "none"
        },
        "Hair": {
            "00": "buzz cut",
            "01": "buzz cut",
            "02": "short, messy hair",
            "03": "short, smooth hair with side part",
            "04": "straight bob cut",
            "05": "wavy bob cut",
            "06": "volumous, slicked-back hair with fade on sides",
            "07": "short mohawk",
            "08": "short, messy mohawk",
            "09": "flat, slicked-back hair with fade on sides",
            "10": "flat, slicked-back hair with fade on sides with man bun",
            "11": "medium length messy waves with middle part",
            "12": "messy hair covering forehead",
            "13": "chest-length hair with side part",
            "14": "ponytail with long bangs",
            "15": "shoulder-length hair with side part covering eye",
            "16": "chest-length hair with middle part",
            "17": "messy chest-length hair pulled back",
            "18": "long ponytail with puffy front",
            "19": "bob cut with side part",
            "20": "bob cut with bangs",
            "21": "long comb-over with fade",
            "22": "medium faux hawk with fade",
            "23": "small afro",
            "24": "small afro with fade",
            "25": "big afro",
            "NULL": "bald"
        },
        "SkinDetails": {
            "Dirt 00": "small specks of dirt on skin",
            "Dirt 01": "large patches of dirt on skin",
            "NULL": "no blemishes on skin",
            "Scars 00": "large scar across face",
            "Scars 01": "large scars across body and face",
            "Tattoo 00": "geometric tattoos on arms, chest, and neck",
            "Tattoo 01": "black-out tattoos on hands and feet",
            "Tattoo 02": "tattoos on arms",
            "Tattoo 03 Female": "tattoo on sternum",
            "Tattoo 03 Male": "tattoo on sternum",
            "Tattoo 04": "curving and dotted tattoos on arms, upper chest, and upper thighs",
            "Tattoo 05": "small symbol tattoos on arms, upper chest, upper thighs, and face",
            "Tattoo 06": "thick stripe tattoos on arms, upper chest, and face",
            "Tattoo 07": "bold tattoos fully covering arms, legs, neck, and upper chest"
        }
    }
}
\end{lstlisting}

Character parameterizer: agent output
\begin{lstlisting}[language=text,numbers=none,basicstyle=\ttfamily,]
The output should be a markdown code snippet formatted in the following schema, including the leading and trailing "```json" and "```":

```json
{
	"cc2d": list[dict]  // Story beats
		{
			"character_options": dict  // Dictionary of chosen options for body parts, body sliders, and RGB values
			"explanation": str  // Explanation for choices
		}
}
```
\end{lstlisting}

image generation for tunic:
\begin{lstlisting}[language=text,numbers=none,basicstyle=\ttfamily,]
tunic with description: {description}, fit the mask, 3/4 perspective, cartoon, 8k
\end{lstlisting}

image generation for pants
\begin{lstlisting}[language=text,numbers=none,basicstyle=\ttfamily,]
pants with this description: {description}, fit the mask,  3/4 perspective front view, cartoon, 8k
\end{lstlisting}

\subsubsection{Sound}

Background sound (looping) description decomposition
\begin{lstlisting}[language=text,numbers=none,basicstyle=\ttfamily,]
You are a sound director that decomposes the an audio description into a composition of several small self-looping audio scripts.
You should read the input JSON and respond with a list in JSON format:

Example input:
```
{
    "desc": "The crackling of the fire and the Village Elder's voice, weaving tales of the forest and its spirits.",
    "people": [{"name": Lily, "desc": a little girl}, {"name": John, "desc": a blacksmith}, {"name": Village Elder, "desc": an old man}]
}
```

Example output:
```json
{
    "audio_scripts": [
        {'type': 'background', 'name': 'fire', 'desc': 'The crackling of the fire in a village', 'duration': 10, 'volume': -35, 'min_pause': 0, 'max_pause': 0},
        {'type': 'foreground', 'name': 'village elder', 'desc': 'An old man is speaking in a village', 'duration': 5, 'volume': -20, 'min_pause': 10, 'max_pause': 30},
    ]
}
```

The meaning of each field in audio scripts is as follows:
- type: either 'background' or 'foreground'. 'background' means the sound is looping seamlessly in the background. 'foreground' means the sound is looping with randomized pause. There should be at least one 'background' sound.
- name: the name of the sound
- desc: the description of the sound. Rewrite the description to replace people's names with its descriptions. For example, replace "John" with "a blacksmith". And describe the sound environment at the end. For example, "in a village".
- duration: the duration of the sound in seconds. Background sounds usually have a duration of 10 seconds. Foreground sounds usually have a duration of 1 ~ 5 seconds.
- volume: the volume of the sound in dB. The volume of background sound is usually around -35 ~ -40 dB. The volume of foreground sound is usually around -20 ~ -30 dB.
- min_pause: for foreground sounds, there should be a pause between each loop. This is the minimum pause in seconds. if the sound only plays once (e.g. the intro music), set this to -1.
- max_pause: this is the maximum pause in seconds. It should be less than 40. if the sound only plays once, set this to -1.

Note:
- Don't use ambiguous words like "forest ambience". Instead, you should break it into specific components like "leaves rustling" and "birds chirping".
- Remove sounds related to narratoring or a single person's speech voice.
\end{lstlisting}

\noindent Sound effects (one-time) description decomposition
\begin{lstlisting}[language=text,numbers=none,basicstyle=\ttfamily,]
You are a sound director that modify an audio description using a given instruction, then decomposes the new description into a composition of several small audio scripts.
You should read the input JSON and respond with a list in JSON format:

Example input:
```
{
    "description": "The sound of pebbles crunching under Sammy's feet as he walks along the river bank.",
    "people": [{"name": Sammy, "desc": a little boy}, {"name": John, "desc": a blacksmith}, {"name": Village Elder, "desc": an old man}]
}
```

Example output:
```json
{
    "duration": 3,
    "audio_scripts": [
        {'name': 'pebbles', 'desc': 'pebbles crunching at a river bank', 'volume': -20},
    ]
}
```

Example input 2:
```
{
    "description": "The hoot of an owl echoes through the woods, followed by the rustling of leaves.",
    "people": [{"name": Sammy, "desc": a little boy}, {"name": John, "desc": a blacksmith}, {"name": Village Elder, "desc": an old man}]
}
```

Example output 2:
```json
{
    "duration": 5,
    "audio_scripts": [
        {'name': 'owl', 'desc': 'the hoot of an owl at a river bank', 'volume': -20},
        {'name': 'leaves rustling', 'desc': 'the rustling of leaves at a river bank', 'volume': -30},
    ]
}
```

The meaning of each field is as follows:
- duration: the duration of all the sound in seconds, usually around 1 ~ 10.
- name: the name of the sound.
- desc: the description of the sound. Simplify the description by removing text that are unrelated to the sound feature. Rewrite the description to replace people's names with its descriptions. For example, replace "John" with "a blacksmith". And describe the sound environment at the end. For example, "in a village".
- volume: the volume of the sound in dB. The volume of sound effects is usually around -40 ~ -20 dB.

Note:
- Don't use ambiguous words like "forest ambience". Instead, you should break it into specific components like "leaves rustling" and "birds chirping".
- Remove sounds related to narratoring or a single person's speech voice.
\end{lstlisting}

\noindent Sound query and rank (first run)
\begin{lstlisting}[language=text,numbers=none,basicstyle=\ttfamily,]
Given a text description about a sound effect, please generate a query string that can be used in a search engine. You should only respond in query string.

# Query format

Queries can be composed of positive tags and negative tags such as:

```
dog bark +city -cat
```
and
```
water +river -sea
```
where positive tags are preceded with "+" and negative ones with "-"

# Your task
{description}
\end{lstlisting}

\noindent Sound query and rank (subsequent run)
\begin{lstlisting}[language=text,numbers=none,basicstyle=\ttfamily,]
You are an assistant for sound effect searching. Based on current search results, please select the best result and refine the search query string to meet the search goal. 

# Query format

Queries can be composed of positive tags and negative tags such as:
```
dog bark +city -cat
```
where positive tags are preceded with "+" and negative ones with "-"

# Note

If no results are found, that means the current query is too strict. 
You should shorten the query down to one or two tags.

# Response format

You should only respond in JSON like this:
```json
{
    "best_index": "the index of the best result",
    "query": "refined query",
}
```

# Search goal
{goal}

# Search history
{history}

# Current query
{query}

# Current results
{results}
\end{lstlisting}

\noindent Sound selection
\begin{lstlisting}[language=text,numbers=none,basicstyle=\ttfamily,]
Given a text description about a sound effect, please choose a sound file from the following list. You should only respond in index number. If none of the files is good enough, respond in -1.
{system_prompt_end}

# Sound description
{description}

# Searched sound files
{results}
\end{lstlisting}

\subsubsection{Speech}

Character simplification
\begin{lstlisting}[language=text,numbers=none,basicstyle=\ttfamily,]
Please simplify the character description using one or two simple adjectives followed by a noun.
You should read the input json and respond with a json that has the same format.

Example input:
```json
[
    {'name': 'Tom', 'age': 6, 'gender': 'male'}, 
    {'name': 'Carol', 'age': 45, 'gender': 'female'},
    {'name': 'Max', 'desc': 'Wearing a black t-shirt and jeans, Max is a young male with short, brown hair and blue eyes'},
]
```

Example output:
```json
[
    {'name': 'Tom', 'desc': 'A young boy'}, 
    {'name': 'Carol', 'desc': 'A middle aged woman'},
    {'name': 'Max', 'desc': 'A young male'},
]
```
\end{lstlisting}

\noindent Prompts for voice fingerprint (using AudioGen)
\begin{lstlisting}[language=text,numbers=none,basicstyle=\ttfamily,]
{character_desc} is speaking clearly and slowly.
\end{lstlisting}

\noindent Prompts for speech reference (using XTTS)
\begin{lstlisting}[language=text,numbers=none,basicstyle=\ttfamily,]
There are a lot of things that I could talk about. But it will probably sound similar to this.
\end{lstlisting}

\noindent Prompts for text-to-speech API.
\begin{lstlisting}[language=text,numbers=none,basicstyle=\ttfamily,]
"{line}" {character} said {emotion_adverb}.
\end{lstlisting}
\end{document}